\documentclass[lettersize,journal]{IEEEtran}
\usepackage{amsmath,amsfonts}
\usepackage{algorithmic}
\usepackage{algorithm}
\usepackage{array}
\usepackage[caption=false,font=normalsize,labelfont=sf,textfont=sf]{subfig}
\usepackage{textcomp}
\usepackage{stfloats}
\usepackage{url}
\usepackage{verbatim}
\usepackage{graphicx}
\usepackage{cite}
\usepackage{booktabs}
\usepackage[table]{xcolor}
\usepackage{multirow}
\usepackage{array}
\usepackage{colortbl}
\usepackage{afterpage}
\usepackage{float}
\usepackage[backref]{hyperref} 
\hypersetup{hidelinks}
\begin{document}

\title{SingingHead: A Large-scale 4D Dataset for \\ Singing Head Animation}

\author{Sijing Wu, Yunhao Li, Weitian Zhang, Jun Jia, Yucheng Zhu, Yichao Yan, \\ Guangtao Zhai,~\IEEEmembership{Senior Member,~IEEE}, and Xiaokang Yang,~\IEEEmembership{Fellow,~IEEE}
}



\maketitle

\begin{abstract}
Singing, as a common facial movement second only to talking, can be regarded as a universal language across ethnicities and cultures, plays an important role in emotional communication, art, and entertainment.
However, it is often overlooked in the field of audio-driven facial animation due to the lack of singing head datasets and the domain gap between singing and talking in rhythm and amplitude. 
To this end, we collect a high-quality large-scale singing head dataset, \textbf{SingingHead}, which consists of more than 27 hours of synchronized singing video, 3D facial motion, singing audio, and background music from 76 individuals and 8 types of music.
Along with the SingingHead dataset, we benchmark existing audio-driven 3D facial animation methods and 2D talking head methods on the singing task. 
Furthermore, we argue that 3D and 2D facial animation tasks can be solved together, and propose a unified singing head animation framework named UniSinger to achieve both singing audio-driven 3D singing head animation and 2D singing portrait video synthesis, which achieves competitive results on both 3D and 2D benchmarks.
Extensive experiments demonstrate the significance of the proposed singing-specific dataset in promoting the development of singing head animation tasks, as well as the promising performance of our unified facial animation framework.
The dataset will be released for research purposes at: \url{https://wsj-sjtu.github.io/SingingHead/}.
\end{abstract}


\section{Introduction}
\IEEEPARstart{A}{udio-driven} facial animation is a challenging task with numerous applications in virtual avatars, extended reality (XR), entertainments, \textit{etc}. Recently, impressive progress has been made in speech-driven talking head animation \cite{zhang2023sadtalker,liu2023moda,liu2024audio,fan2022faceformer,xing2023codetalker,chu2024corrtalk} thanks to the release of a variety of talking head datasets \cite{wu2023mmface4d,zhang2021flow,wang2020mead,fanelli20103,cudeiro2019capture}.
Singing, as another common facial movement, plays a significant role in art and entertainment. People like to express themselves or celebrate through singing on various occasions such as social media and gala, which can be seen as a universal language across ethnicities and cultures. However, how to animate a virtual avatar to sing has been rarely explored.

Singing is quite different from speaking in rhythm, expression, and amplitude (see Fig. \ref{fig:sing_vs_talk}). Therefore, existing methods \cite{fan2022faceformer,xing2023codetalker,thambiraja2023imitator,peng2023selftalk,stan2023facediffuser,wang2021audio2head,gan2023efficient,zhang2023sadtalker,ma2023dreamtalk} trained on talking head datasets encounter difficulties when directly applied to the singing task. 
To solve this problem, it is necessary to collect a singing-specific dataset.
Almost all previous works related to audio-driven facial animation focus on talking, and can be divided into two branches: 
(1) 3D facial animation, \textit{i.e.} generating 3D facial motion given speech audio \cite{cudeiro2019capture,richard2021meshtalk,fan2022faceformer,xing2023codetalker,peng2023emotalk,thambiraja2023imitator,peng2023selftalk,stan2023facediffuser,danvevcek2023emotional,sun2023diffposetalk,zhao2024media2face};
(2) 2D portrait animation, which aims at synthesizing a realistic talking video from a reference portrait image and a driven audio
\cite{yu2020multimodal,prajwal2020lip,zhou2020makelttalk,zhou2021pose,ye2023geneface,shen2023difftalk,sheng2023stochastic,xing2023codetalker,liu2023moda,ma2023dreamtalk,liu2024audio,liu2024osm,zhang2024hierarchical}.
Only a few works focus on singing \cite{iwase2020song2face,liu2023musicface}, however, the datasets they presented are either small in size or of low quality, and none of them provide both 2D portrait videos and 3D facial motions (see Tab.\ref{tab:tab1}), which seriously limits the performance of singing-audio-driven facial animation and the use of these datasets.
The aforementioned limitations severely hinder the development of singing head animation and the flourishing of the automatic virtual avatar singing industry for entertainment.

\begin{table}
\centering
\caption{\textbf{Comparison of Talking/Singing Datasets.} ``Subj.", ``BGM", ``2D", and ``3D" represent subjects, background music, 2D videos, and 3D facial motions, respectively.}
\scalebox{1}{
\begin{tabular}{l|c|cc|ccc}
\toprule

Dataset & Task &Subj.   &Dura.    & BGM   & 2D  & 3D          
\\
\hline
MEAD \cite{wang2020mead} & Talking & 60 & 40h & - &\checkmark  &- \\
HDTF \cite{zhang2021flow} & Talking & 300+ & 15.8h &  - & \checkmark & - \\
\hline  
VOCASET \cite{cudeiro2019capture} & Talking &12   &0.5h   &-   &-   & \checkmark                  
\\
MeshTalk \cite{richard2021meshtalk}  & Talking & 250   &13h   &-   &-   & \checkmark                     
\\
3D-ETF \cite{peng2023emotalk} & Talking &100+   &6.5h   &-   & -   & \checkmark             
\\
\hline 
RAVDESS \cite{livingstone2018ryerson} & Singing & 23  & 2.6h  & - & \checkmark & - \\
Song2Face \cite{iwase2020song2face} & Singing & 7 & 2h & \checkmark & - & \checkmark \\
Musicface \cite{liu2023musicface} & Singing & 6 & 40h & \checkmark & - &  \checkmark  \\
\textbf{Ours} & Singing & 76   & 27h   &\checkmark   &\checkmark   &\checkmark           
\\
\bottomrule
\end{tabular}
}

\label{tab:tab1}
\end{table}

To solve the problem of the scarcity of singing head datasets and few studies on singing head animation tasks, we present \textbf{SingingHead}, a lab-collected large-scale singing head animation dataset, which contains synchronized singing video, 3D facial motion, singing audio, and background music.
Concretely, the design of our dataset considers the following two aspects.
(1) \textbf{Data quality}. We recruit volunteers to collect singing videos in our laboratory rather than collecting in-the-wild videos from the internet to ensure the acquisition of synchronized high-resolution ($3840\times2160$) singing video and dry singing audio, together with an accurate 3D face scan of each volunteer for accurate 3D shape estimation.
The captured videos are cropped to obtain the facial region and serialized into portrait videos with a resolution of $1024\times1024$, which form the 2D video part of our SingingHead dataset.
Moreover, accurate 3D facial motion represented by FLAME \cite{li2017learning} parameters are estimated from the portrait videos with the help of the 3D face scan of each individual.
(2) \textbf{Data diversity.}  
To ensure the diversity of our dataset, we collect singing data from as many as 76 individuals including professional singers and amateur singers. We carefully pick out 8 common music classes, \textit{i.e.} art music, country music, hip-hop, rock music, folk music, rhythm and blues, theater and drama, and march, and ask the volunteers to choose songs they are familiar with in these classes. 
In summary, our SingingHead dataset is the first high-quality dataset that contains both 2D portrait videos and 3D facial motions for singing head animation, which is also comparable in size to the latest 2D and 3D facial animation datasets, as illustrated in Tab. \ref{tab:tab1}.

\begin{figure*}[!t]
\centering
\includegraphics[width=\textwidth]{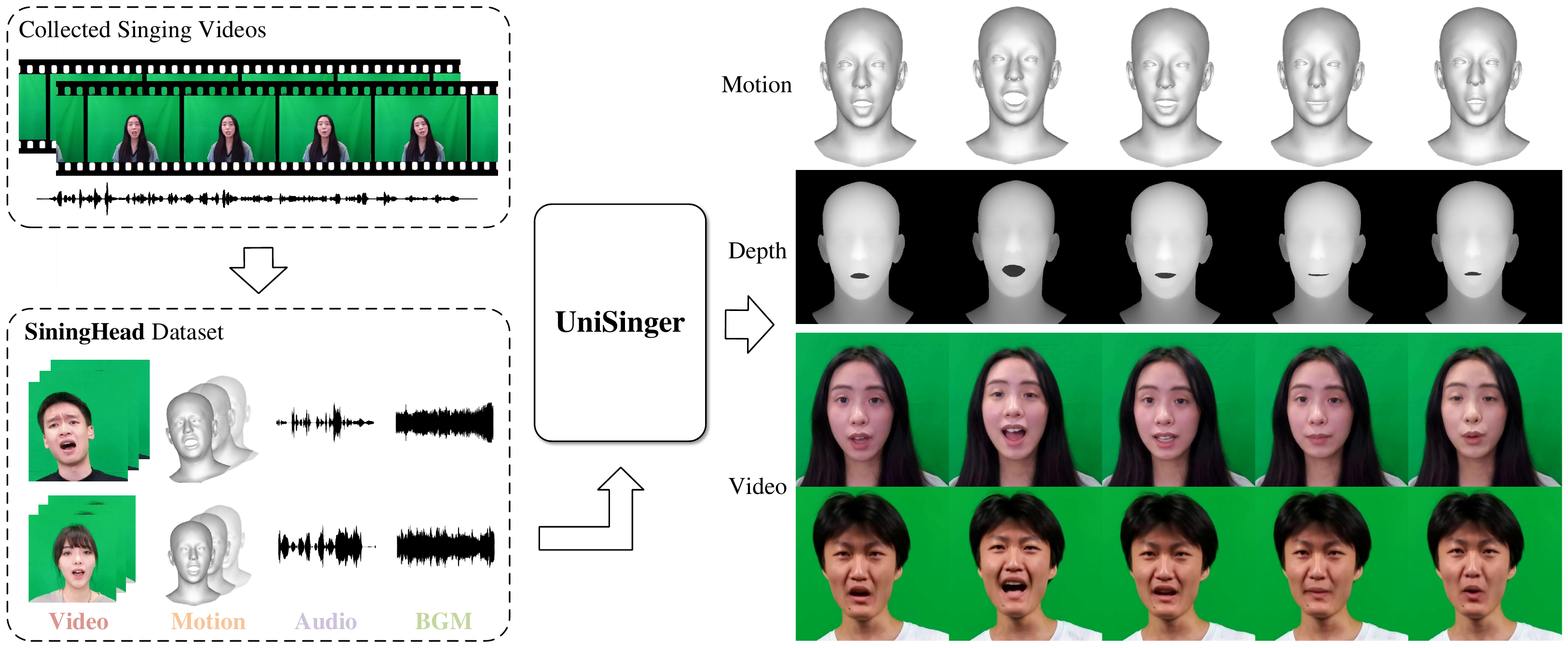}
\caption{We present a new facial animation dataset, \textbf{SingingHead}, which contains more than 27 hours of synchronized singing video, 3D facial motion, singing audio, and background music (BGM) collected from 76 subjects. Along with the SingingHead dataset, we propose a unified framework, UniSinger, to generate both 3D facial motion and 2D singing portrait video according to the input singing audio.}
\label{fig:teaser}
\end{figure*}

Along with the SingingHead dataset, we benchmark existing audio-driven 3D facial animation methods and 2D talking head methods on the singing task, which shows less satisfactory performance. To this end, we further propose a unified framework named \textbf{UniSinger} as a baseline method to solve both 3D singing head animation and 2D singing portrait video synthesis tasks, as shown in Fig.~\ref{fig:teaser}.
For 3D singing head animation, we propose a transformer-based VAE model \cite{vaswani2017attention,kingma2013auto,petrovich2021action} to generate diverse 3D facial motion according to the singing audio, background music, and FLAME \cite{li2017learning} shape parameter.
For 2D singing portrait video synthesis, given a single portrait image and input audio, UniSinger can synthesize singing videos of the person in the reference image which are synchronized with the input audio.
Specifically, we first estimate the shape and camera parameters of the input image. Then the singing audio, background music, and estimated shape parameter are fed into the VAE-based 3D facial animation module to generate 3D facial motion represented by FLAME expression and pose parameters, which is further rendered to depth video using the estimated camera parameters. The depth video together with the input reference image is fed into the face renderer module to generate the final singing portrait video.

To summarize, the main contributions of this paper are:
\begin{itemize}
    \item A large-scale singing head animation dataset, SingingHead, which contains 76 subjects' more than 27 hours of synchronized singing video, 3D facial motion, singing audio, and background music.
    \item Two benchmarks for singing head animation: singing audio-driven 3D singing head animation and 2D singing portrait video synthesis.
    \item A unified framework, UniSinger, which is able to solve both 3D and 2D singing head animation tasks and achieves competitive results on both benchmarks.
\end{itemize}

\section{Related Work}
\label{sec:related}

\subsection{Audio-driven 3D Facial Animation}
In the past few years, audio-driven 3D facial animation \cite{cao2005expressive, taylor2017deep, karras2017audio, zhou2018visemenet} has attracted increasing popularity which can be used to animate various 3D head models \cite{wu2023ganhead,xiang2023flashavatar,chen2023monogaussianavatar,zhu2024dfie3d,jun2016real}. According to the input audio, researchers attempt to obtain animation results with realistic expressions and synchronized mouth movements. 
Benefiting from VOCASET Datset, VOCA \cite{cudeiro2019capture} proposes a speaker-dependent facial animation pipeline that enables generalization across subjects. FaceFormer \cite{fan2022faceformer} introduces transformer \cite{vaswani2017attention} to capture long-term audio context and achieve temporary consistent animation performance. However, both of these two works lack attention to upper-face motions. To alleviate this problem, Meshtalk \cite{richard2021meshtalk} learns a categorical latent space via a dedicated cross-modality loss. 
Subsequently, numerous works \cite{xing2023codetalker,peng2023emotalk,thambiraja2023imitator,stan2023facediffuser,danvevcek2023emotional,zhong2024expclip,aneja2023facetalk,sun2023diffposetalk,zhao2024media2face,chu2024corrtalk} are proposed to further improve the realism.
Codetalker \cite{xing2023codetalker} utilizes a discrete motion prior based on VQ-VAE \cite{van2017neural} to avoid over-smoothed animation results. FaceDiffuser \cite{stan2023facediffuser} and DiffPoseTalk \cite{sun2023diffposetalk} introduce diffusion model \cite{ho2020denoising} to achieve diverse and realistic 3D facial motion generation, while DiffPoseTalk further considers the head pose generation.
Meanwhile, Emotalk \cite{peng2023emotalk}, EMOTE \cite{danvevcek2023emotional} and ExpCLIP \cite{zhong2024expclip} explore the emotional speech-driven 3D facial animation.
However, all these methods focus on talking and are trained on talking datasets, leading to poor performance on the singing task. 
To this end, we collect the SingingHead dataset and benchmark existing 3D facial animation methods using the 3D facial motion of the dataset. An efficient baseline method that utilizes the background music is also proposed and benchmarked.
Recently, Media2Face \cite{zhao2024media2face} shows some singing ability, however, it is not specialized for the singing task and is not open source. In contrast, we first propose the 4D singing head animation dataset and explore the 3D singing head animation task \cite{wu2023singinghead}.


\subsection{Audio-driven 2D Portrait Animation}
2D portrait animation aims to generate lifelike videos that synchronized with the input audio \cite{zhou2012image,son2017lip,edwards2016jali, zhu2018arbitrary,zhou2020makelttalk,wang2021audio2head,song2022audio,zhang2023sadtalker,liu2023moda,gan2023efficient,shen2023difftalk,zhang2023metaportrait,peng2024synctalk,ma2023dreamtalk,tian2024emo,xu2024vasa,xu2024hallo}. Through the integration of a pre-trained Lip-Sync expert \cite{Chung2016lip_sync}, Wav2Lip \cite{prajwal2020lip} enhances synchronization accuracy of the generated results.
MekeItTalk \cite{zhou2020makelttalk} improves the realism via a speaker-aware audio encoder that enables personalized head pose movement. To enable natural head pose movement, PC-AVS \cite{zhou2021pose} proposes a low-dimensional pose code to disentangle and control head poses. With the development in generation and reconstruction tasks, many methods \cite{ye2023geneface, shen2023difftalk, guo2021ad, yao2022dfa, liu2022semantic, stypulkowski2023diffused} introduce Neural Radiance fields \cite{mildenhall2021nerf} or diffusion model \cite{ho2020denoising} to refine the quality of the generated videos. However, these methods typically suffer from less efficient generation processes due to the time-consuming volume rendering or diffusion sampling procedures. MODA \cite{liu2023moda} proposes a dual-attention module to estimate dense facial landmarks of deterministic and probabilistic movements to generate natural and detailed results. In recent years, some works \cite{ji2022eamm,gan2023efficient,tan2024flowvqtalker} have explored the emotional talking head animation to achieve more realistic results.
However, these methods mainly focus on speech audio, while another common type of audio, singing, remains under-explored. Since singing audio has a different rhythm compared to speech audio, and the expressions during singing are more exaggerated, it is challenging for current methods to directly generalize to the singing task. Therefore, we propose a large-scale singing head dataset for the development of singing head animation.
Although some latest works \cite{tian2024emo,xu2024vasa,xu2024hallo} show some singing demos, they do not delve into the singing problem in depth. We are the first to study the singing head animation task based on our singing-specific dataset \cite{wu2023singinghead}.

\begin{figure*}[t]
\centering
\includegraphics[width=\textwidth]{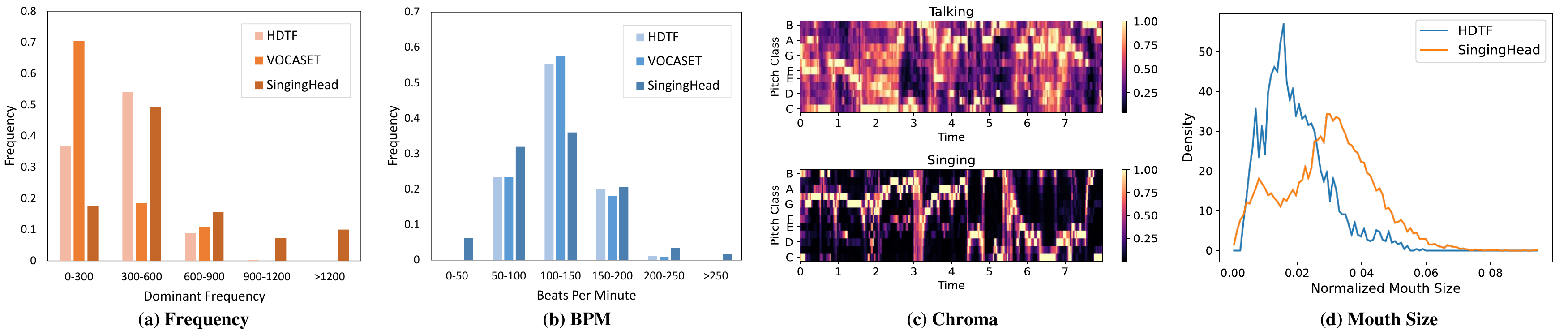}
\caption{\textbf{Comparison between talking and singing.} (a) and (b) are the bar charts of the dominant frequency distribution and BPM distribution of audio, respectively. Our singing dataset has a wider range of dominant frequency and BPM distribution compared to talking datasets. (c) shows the chromagrams of a talking audio and a singing audio, which are very different and show that singing has more regular and stable pitch patterns as well as organized energy distribution. (d) is the distribution of mouth opening size calculated from facial landmarks, which shows that the mouth size is typically larger when singing compared to talking. All these demonstrate that singing is quite different from talking, and therefore needs to be studied separately.}
\label{fig:sing_vs_talk}
\end{figure*}

\subsection{2D and 3D Facial Animation Datasets} 
Facial Animation Datasets play an important role in facial animation research, which can be divided into 2D and 3D datasets.
2D datasets \cite{livingstone2018ryerson, wang2020mead, prajwal2020lip, nagrani2017voxceleb, cao2014crema, jackson2014surrey, zhang2021flow} focus on collecting more diverse and higher quality photorealistic data. LRW dataset \cite{prajwal2020lip} contains 173 hours of low-resolution videos. To enlarge the data scale, the VoxCeleb dataset \cite{nagrani2017voxceleb} collected more than 2k hours of videos from YouTube. In addition, several works \cite{cao2014crema, livingstone2018ryerson,wang2020mead} explore collecting emotional talking head data with emotion labels. The SAVEE \cite{jackson2014surrey} dataset is the first emotional dataset with only 4 actors. MEAD \cite{wang2020mead} dataset builds a high-resolution dataset with various video clips and viewpoints, covering a large number of emotional audio-visual corpus. 
Numerous 3D datasets \cite{cudeiro2019capture, richard2021meshtalk, fanelli20103,wu2023mmface4d} have also been proposed for audio-driven 3D facial animation. MeshTalk dataset \cite{richard2021meshtalk} contains detailed meshes topology and proposes a convolution neural network-based method. VOCASET dataset \cite{cudeiro2019capture} constructs dataset by registering meshes to the FLAME \cite{li2017learning} topology. However, these datasets are scale-limited. To this end, MMFace4D \cite{wu2023mmface4d} collects a larger dataset consisting of 36 hours of data from 431 subjects. Recently, some works \cite{peng2023emotalk,danvevcek2023emotional,sun2023diffposetalk} have built 3D facial animation datasets from 2D portrait videos using 3D face reconstruction technologies to scale up the 3D dataset.
However, all these datasets are collected for general talking and lack the generalization ability for singing tasks. 
To this end, a few works \cite{iwase2020song2face,liu2023musicface} try to solve the singing task by collecting singing-specific datasets.
Song2Face dataset \cite{iwase2020song2face} collects a 2-hour dataset with background music and 3D blendshapes from 7 singers. MusicFace \cite{liu2023musicface} constructs a 3D dataset with only 6 singers by collecting the audio and intermediate videos separately. However, the existing two singing datasets are either small in size or of low quality, and neither provides 2D portrait singing videos, which limits their applications.
Different from them, our SingingHead dataset is the first singing head dataset containing both 2D portrait videos and accurate 3D facial motions, with the greatest diversity (76 singers) and large scale (27 hours).

\section{SingingHead Dataset}

\begin{figure*}[h]
\centering
\includegraphics[width=\linewidth]{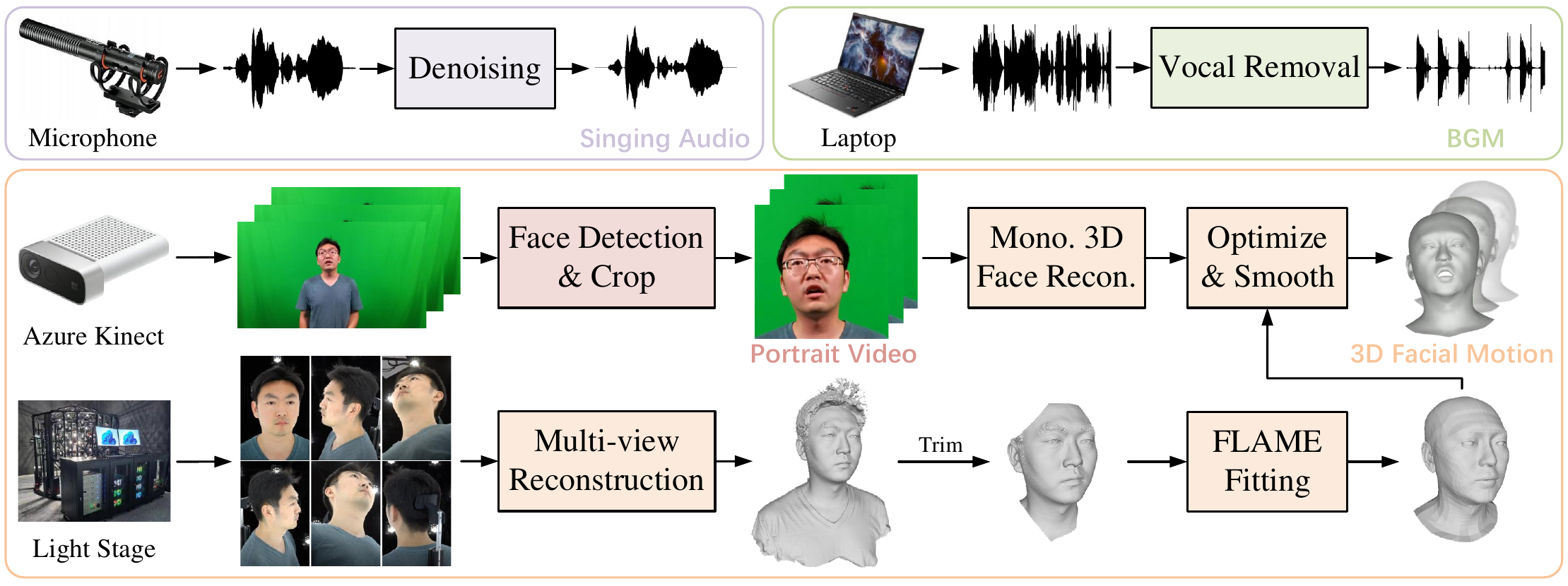}
\caption{\textbf{Data collection pipeline.} We use microphone, laptop, Azure Kinect camera, and light stage to collect raw data including singing audio, original songs downloaded from the Internet, singing video, and 3D head scan. Then, we process the raw data to obtain the cleaned singing audio, background music, portrait video, and 3D facial motion, which make up our SingingHead dataset.}
\label{fig:dataset}
\end{figure*}

Considering the lack of high-quality facial animation datasets specifically focused on singing, we collect SingingHead, a high-quality singing head animation dataset that contains both portrait video and accurate 3D facial motion with synchronized singing audio and background music.
The dataset collection pipeline is shown in Fig. \ref{fig:dataset}.

\subsection{Data Acquisition}
\begin{figure*}[t]
\centering
\includegraphics[width=\linewidth]{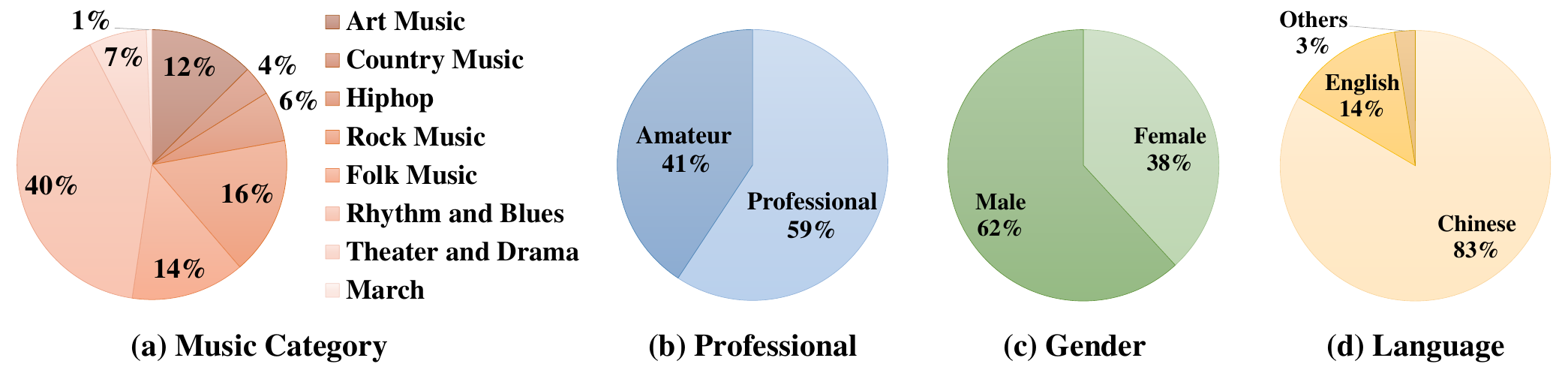}
\caption{\textbf{Data Statistics.} We show the pie charts of (a) the music category distribution, (b) the professional distribution of volunteers, (c) the gender distribution of volunteers, and (d) the language distribution, respectively.}
\label{fig:statistics}
\end{figure*}

\noindent\textbf{Design policy.} 
We design the collection of SingingHead dataset by considering the following three points. 
(1) \textbf{Variety of songs}. For the diversity of the singing dataset, we carefully selected 8 common music categories, \textit{i.e.}, art music, country music, hip-hop, rock music, folk music, rhythm and blues, theater and drama, march, and asked the volunteers to choose 5 to 10 songs they were familiar with and fall to these categories. The pie chart in Fig. \ref{fig:statistics} (a) illustrates the categories and distributions of the songs we have finally collected.
(2) \textbf{Professionalism of volunteers}. Different from talking, singing is a form of art that requires practice, which means that there is a great difference in singing skills between those who practice singing frequently and those who sing less. To this end, we select 50 professional singers and 26 amateurs as our volunteers, and recorded a total of 265 songs performed by professional volunteers and 182 songs performed by amateur volunteers, as illustrated in Fig. \ref{fig:statistics} (b).
(3) \textbf{Acquisition of dry sound}. In normal circumstances, singers need to listen to the accompaniment or the original song when singing. In order to record the dry sound of singing, we equipped the volunteers with small Bluetooth earphones to play the accompaniment, and then recorded their voices via another microphone.

\noindent\textbf{Equipment and process.} 
We leverage Azure Kinect Camera to capture RGB video at 30 fps with a resolution of $3840\times2160$. We record audio using the COMICA VM20 microphone. We also download original songs from the Internet and use software to remove the vocals to obtain pure background music. 
To supplement our sequential data with high-quality shape information, we further use light stage cameras to synchronously capture photos of the volunteers up to 6$K$ resolution in nature expression from 51 different views and then reconstruct fine-gained 3D scans using Metashape~\cite{metashape}. Concretely, we played the original song (in Bluetooth earphones), and subtitles (on the screen), and recorded the volunteers' singing video and audio simultaneously. It is worth mentioning that all these four operations are fully synchronized, and we record song by song. After a volunteer finishes all his songs, we capture his multi-view photos using the light stage.

\subsection{Data Processing}
The raw data includes RGB videos, singing dry sounds, original songs, and 3D scans. We processed the raw data as illustrated in Fig. \ref{fig:dataset} to get our final dataset SinginHead, which consists of singing audio, background music, portrait video, and 3D facial motion sequence.

\begin{figure}
\centering
\includegraphics[width=\linewidth]{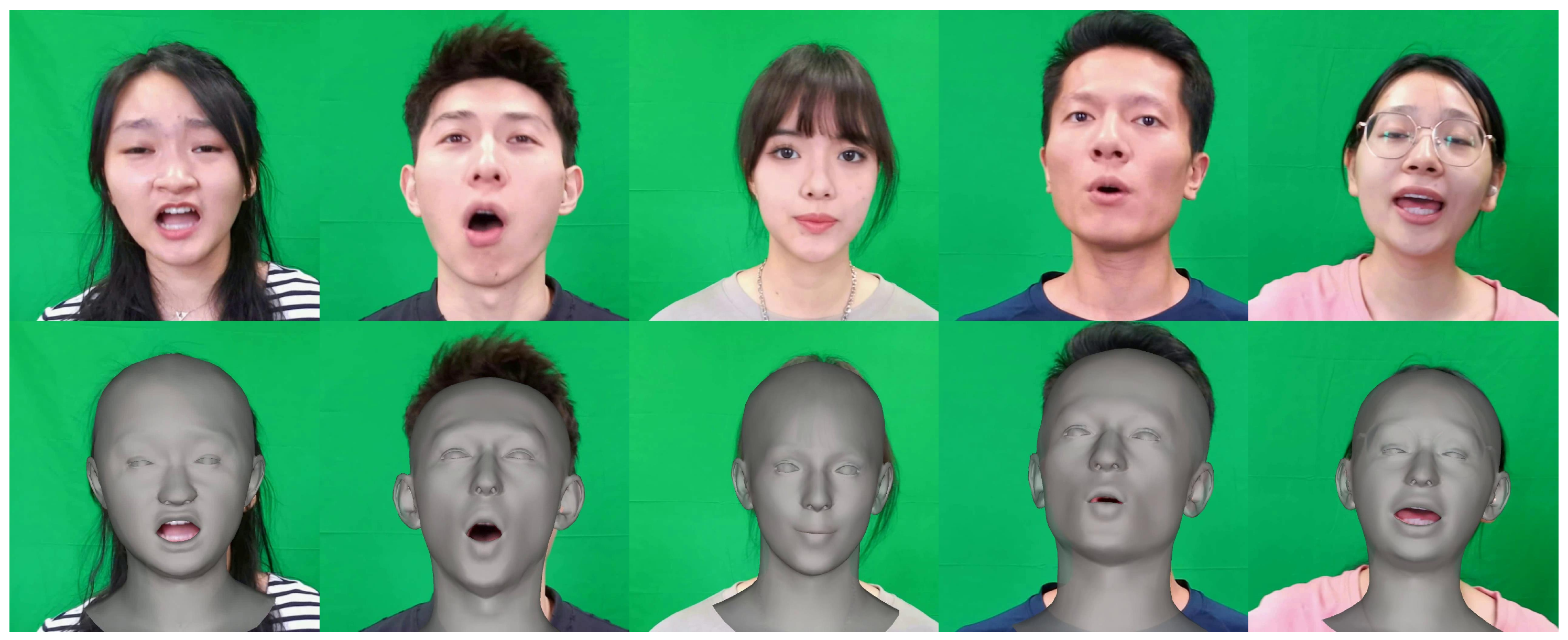}
\caption{\textbf{Visualization of SingingHead dataset.} We show some key frames from singing portrait videos and the corresponding 3D facial motions.}
\label{fig:show}
\end{figure}

\noindent\textbf{Video cropping and serialization.}
Since the video we collected has a wide field of view, including the entire upper body, and the volunteers routinely wiggled as they recorded the song, it is necessary to crop out the facial region and serialize the long recording to ensure the face remains in a stable bounding box. To this end, we first detect the bounding box of the facial region using MTCNN \cite{zhang2016joint}, then leave a margin around the bounding box to determine the crop box and crop the original video into square portrait video, finally, the portrait video is resized to $1024\times1024$ as shown in Fig. \ref{fig:show}. During the cropping process, we use Dlib \cite{king2009dlib} to detect the 68 landmarks every 6 frames, and calculate whether the landmarks on the edge are too close to the bounding box, if so, we cut the video at this frame, and start a new video sequence with a newly detected bounding box.

\noindent\textbf{3D facial motion reconstruction.}
We utilize the cropped portrait videos and the corresponding 3D scans to obtain accurate 3D singing facial motion sequences. Specifically, the 3D facial motion is represented using the expression and pose parameters of the commonly used parametric head model FLAME \cite{li2017learning}.
To estimate accurate FLAME parameters of the singing head, we first estimate the shape, expression, pose, and camera parameters separately for each frame in the portrait video with a SOTA monocular 3D face reconstruction method EMOCA v2 \cite{danvevcek2022emoca, DECA:Siggraph2021, filntisis2022visual}, and average the estimated shape parameters of all frames of a person as his/her shape parameter. Considering that estimating 3D shape from the 2D image is an ill-posed problem, and it is hard to obtain very accurate results, we further use the collected fine-gained 3D scans to fit a more accurate FLAME model by minimizing the 3D landmark loss and mesh-to-scan distance with the shape parameter initialized as the monocular estimated shape.
Considering the FLAME parameters are estimated frame-by-frame, we utilize Box Plot to detect the outliers and replace them with the values of neighboring points to avoid discontinuous or incorrect motions. Concretely, we first calculate the average value of the estimated parameters over time, and then calculate Euclidean distance of each value to the corresponding average value and replace the outliers whose distances are outside the interval $[Q_1-s(Q_3-Q_1), Q_3+s(Q_3-Q_1)]$, where $Q_1$ and $Q_3$ are the first and third quartile respectively, $s$ is the scale factor.
Then, we smooth the detected 68 landmarks for the next optimization process by minimizing $||(\boldsymbol{lmk}_{2:T} -\boldsymbol{lmk}_{1:T-1} )||_2^2$ using Adam optimizer \cite{kingma2014adam}.
Finally, we use the fitted FLAME shape as the final shape parameter, and reproject the estimated FLAME head back to the original portrait video to optimize the global orientation, neck pose, and camera parameters by minimizing:
\begin{equation}
\mathcal{L} = \lambda_{l}\mathcal{L}_{lmk} + \lambda_s\mathcal{L}_s + \lambda_r\mathcal{L}_{reg},
\label{eq:1}
\end{equation}
where $\mathcal{L}_{lmk}$, $\mathcal{L}_{s}$ and $\mathcal{L}_{reg}$ denotes landmark reprojection loss, smoothing loss, and regularization term, respectively. $\lambda_l$, $\lambda_s$ and $\lambda_r$ are the weight coefficients. 
Specifically, landmark reprojection loss $\mathcal{L}_{lmk}$ is calculated by:
\begin{equation}
\mathcal{L}_{lmk} = \left \| (\boldsymbol{\hat{lmk}} -\boldsymbol{lmk} ) \right \|_2^2,
\label{eq:2}
\end{equation}
where $\boldsymbol{\hat{lmk}}$ and $\boldsymbol{lmk}$ represent ground truth and reprojected 2D landmarks respectively. Smoothing loss $\mathcal{L}_{s}$ is the sum of the smoothing losses of global orientation, neck pose, camera and bounding box parameters with same formulation: $||(\boldsymbol{X}_{2:T} -\boldsymbol{X}_{1:T-1} )||_2^2$ where $\boldsymbol{X}$ represent any one of these variables.
The regularization term $\mathcal{L}_{reg} = ||\boldsymbol{\theta_g}||_2^2 + ||\boldsymbol{\theta_n}||_2^2$, where $\boldsymbol{\theta_g} \in \mathbb{R}^{3}$ represents the global orientation and $\boldsymbol{\theta_n} \in \mathbb{R}^{3}$ is the FLAME neck pose parameter.

After these steps, we can obtain accurate and smooth 3D singing facial motions represented by 59-dimensional FLAME \cite{li2017learning} parameters (\textit{i.e.,} 3 dimensions for global orientation, 50 dimensions for expression, 3 dimensions for jaw pose and 3 dimensions for neck pose). Besides, all the subjects have their 100-dimensional FLAME shape parameter. Some key frames of our 3D facial motions are shown in the second row of Fig. \ref{fig:show}, demonstrating the high precision of the reconstructed 3D facial motions.

\subsection{Dataset Analysis}
\noindent\textbf{Data statistics.} 
According to the aforementioned pipeline, we record a total of 448 raw video sequences with synchronized singing audio from 76 subjects. Each sequence comes from a complete song, ranging from 27 to 457 seconds. Fig. \ref{fig:statistics} shows pie charts of the music category distribution and language distribution of these recordings, as well as the distribution of whether the subjects are professional and their gender.
Then, portrait videos are cut into smaller segments to keep the facial region within the cropped bounding box, resulting in 10223 video sequences. The singing audio, background music, and 3D facial motion are then segmented according to the video segments. Sequences less than 3s are all removed. Therefore, the SingingHead dataset consists of 1952 synchronized singing videos, 3D facial motions, singing audios, and background music ranging from 3 to 432 seconds.
For convenience, we also cut these sequences into equal-length 8s segments for benchmarking and training our method, resulting in a total of 12196 8s sequences.

\noindent\textbf{Differences from talking head datasets.} 
As shown in Fig. \ref{fig:sing_vs_talk}, our singing dataset is quite different from existing talking datasets both in terms of audio features and facial movements, so it is necessary to collect the singing-specific dataset.
To be specific, singing audio has a wider range of frequency and BPM (Beats Per Minute) due to the diversity of songs and their broader range of tones and rhythms. 
In talking datasets, people usually speak at moderate speaking rate and intonation, so that the frequencies and BPMs of talking audio are concentrated in the lower frequency range and middle range, respectively.
Different from talking datasets, we ask the volunteers to sing a variety of songs including fast songs, slow songs, bass songs, and treble songs, which results in a much wider range of frequency and BPM distribution of the singing audio in our SingingHead dataset.
Moreover, people often exhibit more exaggerated expressions when singing compared to talking for artistic performance and emotional communication. For instance, the mouth opening size is typically larger when singing compared to talking as shown in Fig. \ref{fig:sing_vs_talk} (d). Besides, there tend to be more head movements when singing than when talking. Therefore, it is essential for models to learn these movements from singing-specific datasets to achieve vivid and realistic singing head animation.


\section{Method}

\begin{figure}
\centering
\includegraphics[width=\linewidth]{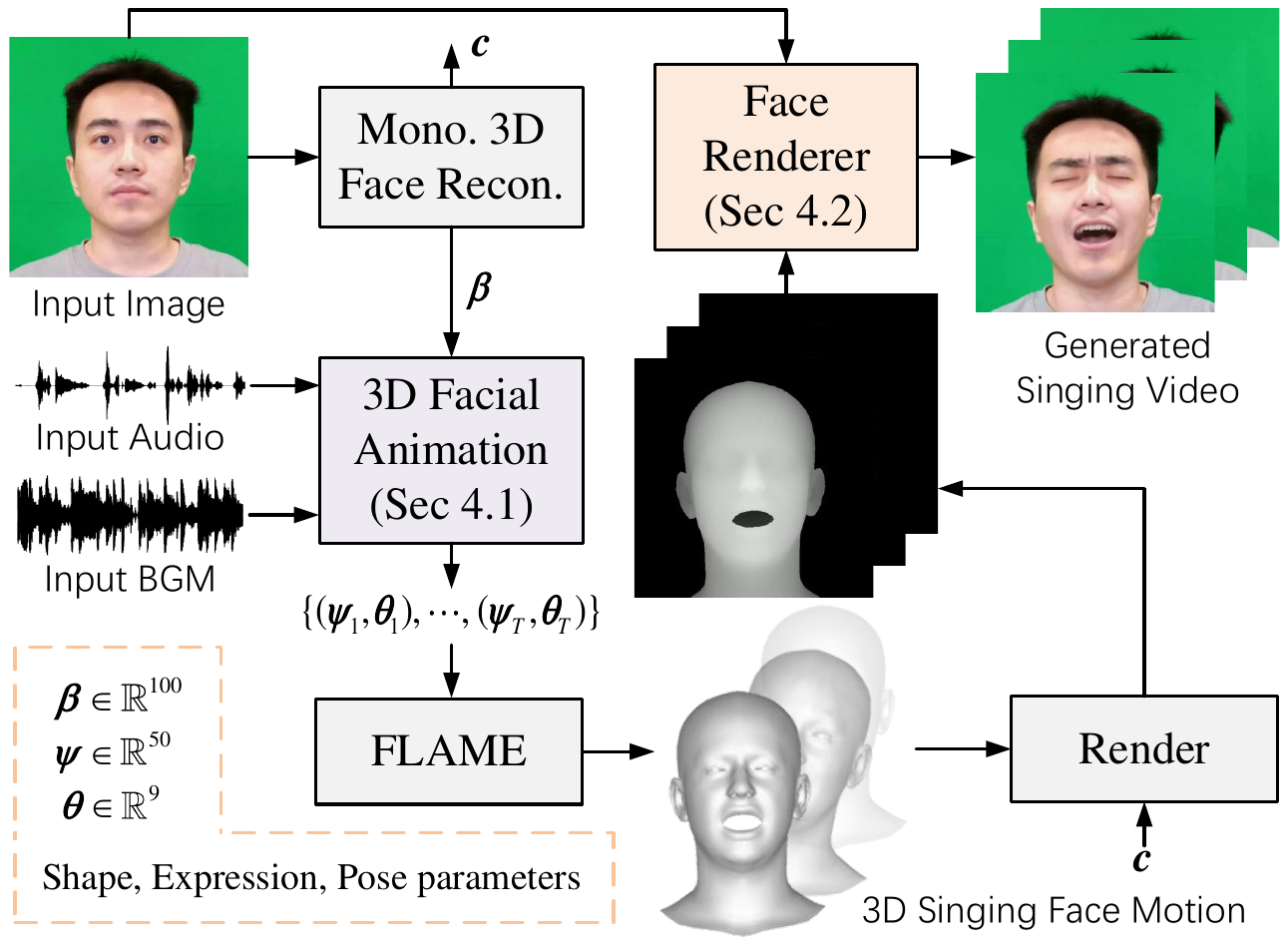}
\caption{\textbf{Overview of the UniSinger framework.} Given singing audio, background music, and shape parameter, the 3D facial animation module can generate diverse singing facial motions synchronized with the input singing signal. The generated 3D facial motions can be further rendered into 2D singing videos through the face renderer module.}
\label{fig:overview}
\end{figure}

\begin{figure*}[h]
\centering
\includegraphics[width=\textwidth]{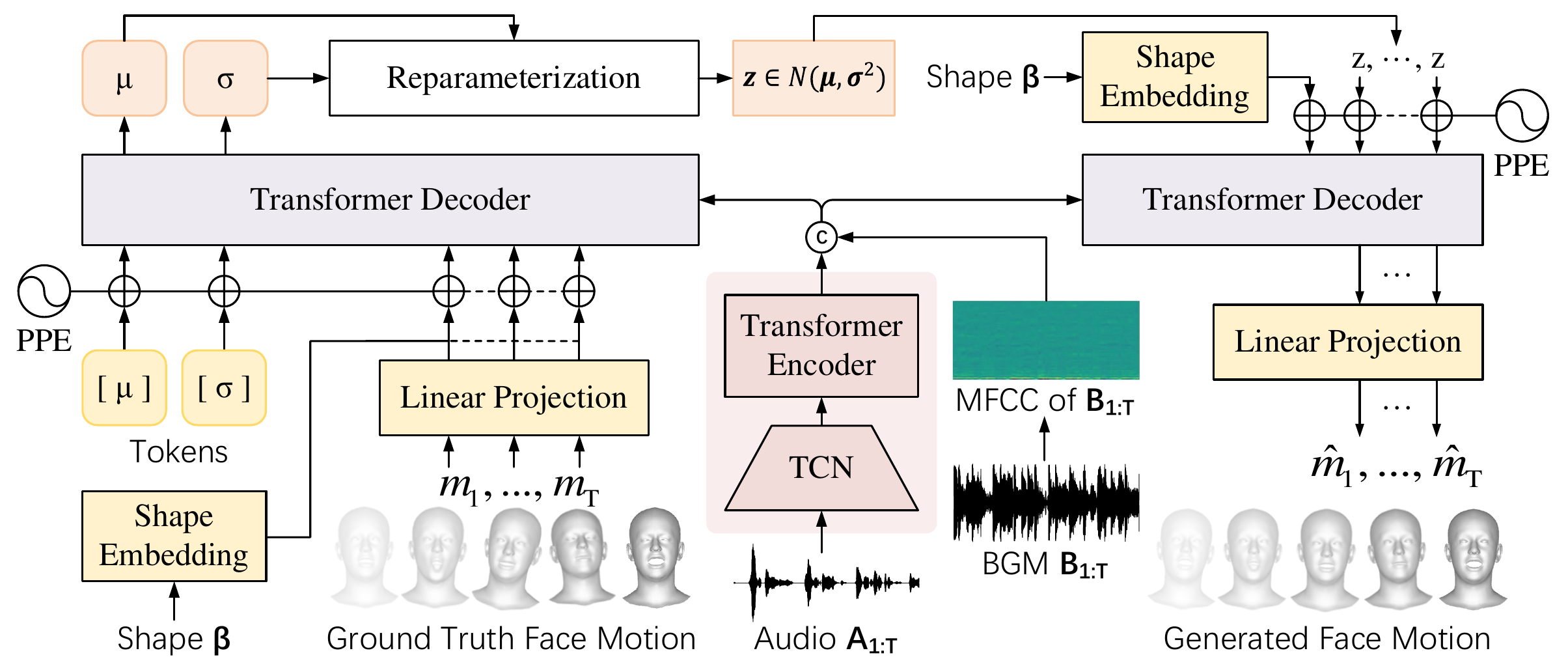}
\caption{\textbf{VAE-based 3D facial animation.} 
Given the 3D facial motion sequence $\{\boldsymbol{m}_1,...,\boldsymbol{m}_T\}$, FLAME shape parameter $\boldsymbol{\beta}$, singing audio $\boldsymbol{A}_{1:T}$ and background music $\boldsymbol{B}_{1:T}$, the transformer-based VAE encoder outputs the distribution parameters $\boldsymbol{\mu}$ and $\boldsymbol{\sigma}$. By sampling from the distribution, latent vector $\boldsymbol{z}$ is obtained and fed into the transformer-based VAE decoder together with the shape parameter and input audio, and output the generated facial motion sequence $\{\boldsymbol{\hat{m}}_1,...,\boldsymbol{\hat{m}}_T\}$. During inference time, we can generate diverse facial motion sequences according to the input audio and shape parameter by sampling the learned distribution via the VAE decoder. 
}
\label{fig:method}
\end{figure*}


Along with the SingingHead dataset, we argue that 3D and 2D facial animation tasks can be solved together in one pipeline. To this end, we propose a unified framework called \textbf{UniSinger} to generate diverse 3D facial motion according to the input singing audio (Sec.\ref{sec:4.1}), which can be further rendered into 2D portrait singing videos (Sec.\ref{sec:4.2}). At inference time, given a single image, singing audio, and background music, we first reconstruct the FLAME \cite{li2017learning} shape parameters $\boldsymbol{\beta}$ and camera parameter $\boldsymbol{c}$ from the input image through a monocular 3D reconstruction method \cite{danvevcek2022emoca}. Then, we feed the audio and shape parameter into the 3D facial animation module to get the 3D singing motion represented by a sequence of FLAME expression $\boldsymbol{\psi}$ and pose $\boldsymbol{\theta}$ parameters. The generated 3D facial can be directly used for virtual avatar animation, or further rendered to depth video and fed into our face renderer together with the single input image to get a 2D portrait singing video.
The overview of our pipeline is illustrated in Fig. \ref{fig:overview}.

\subsection{VAE-based 3D Facial Animation}
\label{sec:4.1}
Given singing audio $\boldsymbol{A}_{1:T} = \{\boldsymbol{a}_1, ...,\boldsymbol{a}_T \}$, background music $\boldsymbol{B}_{1:T} = \{\boldsymbol{b}_1, ...,\boldsymbol{b}_T \}$ and FLAME shape parameter $\boldsymbol{\beta} \in \mathbb{R}^{100}$, our VAE-based 3D facial animation module can generate diverse 3D facial motion $\boldsymbol{M}_{1:T} = \{\boldsymbol{m}_1, ...,\boldsymbol{m}_T \}$, where $\boldsymbol{m}_i=(\boldsymbol{\psi}_i, \boldsymbol{\theta}_i)$ denotes the motion of $i$-{th} frame which is represented by FLAME expression $\boldsymbol{\psi} \in \mathbb{R}^{50} $ and pose $\boldsymbol{\theta} \in \mathbb{R}^{50} $ parameters. 
Concretely, we utilize a conditional variational autoencoder (CVAE) model \cite{sohn2015learning} inspired by \cite{petrovich2021action} to construct our 3D facial animation module, which is built on transformer layers \cite{vaswani2017attention} and conditioned on shape and audio embeddings (Fig. \ref{fig:method}).

\noindent\textbf{Transformer-based VAE encoder.}
The VAE encoder takes a facial motion sequence and the corresponding shape parameter and singing audio as input, and outputs the distribution parameters $\boldsymbol{\mu}$ and $\boldsymbol{\sigma}$ of the singing facial motion space. Then a ${d}$-dimensional latent vector $\boldsymbol{z} \in \mathbb{R}^{d}$ is sampled from the Gaussian distribution using reparameterization \cite{kingma2013auto}, which will be fed to the VAE decoder.
To be specific, the facial motion sequence is first projected into a ${d}$-dimensional space via a linear projection layer.
To model the style of facial motion of different people during singing, the FLAME shape parameter, which varies from person to person, is embedded to a ${d}$-dimensional feature vector via the shape embedding layer. 
As for the input singing audio, we build an audio encoder using the structure of a SOTA speech model, wav2vec 2.0 \cite{baevski2020wav2vec} followed by a linear projection layer to project the output audio feature to the $d$ dimension following \cite{fan2022faceformer,xing2023codetalker}, and initialize our audio encoder using the pre-trained wav2vec 2.0 weights.
Considering that background music mainly provides rhythmic and beat information instead of phonemes, we utilize its low-level MFCC features \cite{sahidullah2012design,yi2023generating}. Then, a linear projection layer is used to project the 64-dimensional MFCC features to the $d/2$ dimension, which is then concatenated with the singing audio feature to obtain the entire audio feature. Finally, the entire audio feature is projected to $d$ dimension using another linear projection layer.
Since the number of tokens remains unchanged across multiple transformer layers, we add two additional learnable tokens $[\mu]$ and $[\sigma]$ to extract the distribution parameters $\boldsymbol{\mu}$ and $\boldsymbol{\sigma}$ inspired by \cite{dosovitskiy2020image, petrovich2021action}.
After these steps, all feature dimensions are unified to a ${d}$-dimensional format. We add shape embedding to the projected motion sequence, and then feed them together with the learnable tokens into a transformer decoder \cite{vaswani2017attention} after applying a periodic positional encoding (PPE) \cite{fan2022faceformer} to all the input tokens. 
At the same time, the entire audio feature is fed into the transformer decoder through cross-attention layers with a diagonal attention mask to align the audio with the facial motion sequence.
Finally, the first two tokens of the transformer outputs are picked out as the distribution parameters $\boldsymbol{\mu}$ and $\boldsymbol{\sigma}$.

\noindent\textbf{Transformer-based VAE decoder.}
Given $T$ latent vectors $\{\boldsymbol{z}, ..., \boldsymbol{z}\}$, together with shape parameter and audio, the VAE decoder generates a $T$-frames 3D facial motion sequence according to the input audio.
Concretely, the shape parameter and audio are embedded in the same way as in the VAE encoder, and PPE \cite{fan2022faceformer} is applied to the shape embedding and $T$ latent vectors. Then, the output of PPE is fed into another transformer decoder together with the entire audio feature through cross-attention layers similar to the VAE encoder. The transformer outputs without the first toke is finally fed to a linear projection layer to get the generated 3D facial motion $\boldsymbol{\hat{M}}_{1:T} = \{\boldsymbol{\hat{m}}_1, ...,\boldsymbol{\hat{m}}_T \}$.

\noindent\textbf{Training and inference.}
We train the VAE-based 3D facial animation module using the audio, shape parameter, and facial motion pairs in our SingingHead Dataset. 
We use the loss function $\mathcal{L} = 
\lambda_{k}\mathcal{L}_{kl} + \lambda_{r}\mathcal{L}_{rc} + \lambda_{v}\mathcal{L}_{vel} + \lambda_{m}\mathcal{L}_{mesh}$ to train our model, where $\mathcal{L}_{vel}$, $\mathcal{L}_{rc}$, $\mathcal{L}_{kl}$, $\mathcal{L}_{mesh}$ stands for KL loss, reconstruction loss, velocity loss and mesh vertices loss, respectively. 
Specifically, KL loss $\mathcal{L}_{kl}$ aims at regularizing the latent distribution to a standard Gaussian distribution, which is the same as the standard VAE \cite{kingma2013auto}. Reconstruction loss measures the distance between the generated facial motion $\boldsymbol{\hat{M}}_{1:T}$ and the ground truth facial motion $\boldsymbol{M}_{1:T}$, which is written as:
\begin{equation}
\mathcal{L}_{re} = || \boldsymbol{\hat{M}}_{1:T}  - \boldsymbol{M}_{1:T} ||_2^2.
\end{equation}
Velocity loss measures the velocity difference between $\boldsymbol{\hat{M}}_{1:T}$ and $\boldsymbol{M}_{1:T}$, which is written as:
\begin{equation}
\mathcal{L}_{vel} = ||(\boldsymbol{\hat{M}}_{2:T}-\boldsymbol{\hat{M}}_{1:T-1})- (\boldsymbol{M}_{2:T} - \boldsymbol{M}_{1:T-1})||_2^2.
\end{equation}
Mesh vertices loss measures the L2 distance between the vertices of the face region from the generated and ground truth mesh sequences. The meshes are obtained by applying the FLAME model \cite{li2017learning} to the generated or ground truth FLAME parameters.

During inference time, only the VAE decoder is used. Concretely, given input audio, FLAME shape parameter, and the latent vector $\boldsymbol{z}$ randomly sampled from Gaussian distribution, our VAE decoder can generate realistic and accurate 3D singing facial motion synchronized with the input audio. Moreover, our model can generate diverse facial motions with the different sampling results of $\boldsymbol{z}$.

\begin{figure}
\centering
\includegraphics[width=\linewidth]{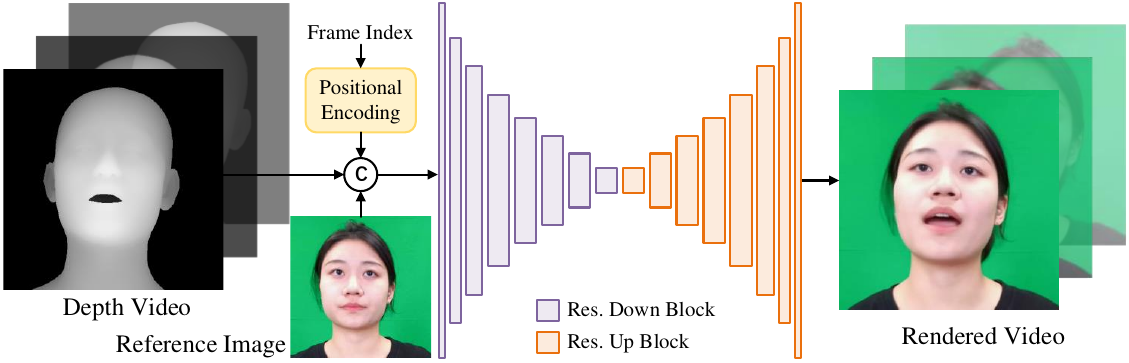}
\caption{\textbf{Structure of the Face Renderer.} Given a depth video that represents a singing motion and a reference image, the face renderer module can generate a singing portrait video through a UNet-like convolutional neural network with residual blocks.}
\label{fig:render}
\end{figure}

\subsection{Face Renderer}
\label{sec:4.2}
As shown in Fig. \ref{fig:render}, the face renderer module takes the FLAME depth image sequence and a reference image as input, rendering it to a portrait singing video with the same identity as the reference image. 
Specifically, the face renderer module is actually a generator constructed based on a convolutional neural network with skip connections following \cite{lu2021live,liu2023moda}. In detail, the encoder of the generator contains 8 layers with resolutions of $(256^2, 128^2, 64^2, 32^2, 16^2, 8^2, 4^2, 2^2)$ and corresponding feature channel numbers of $(64, 128, 256, 512, 512, 512, 512, 512)$. The structure of the decoder is symmetric to the encoder, with upsampling layers.
Inspired by \cite{liu2023moda}, temporally positional embedding (TPE) is concatenated with the input depth image and reference image and then fed into the generator to improve temporal stabilization.
The generator is trained together with a PatchGAN \cite{isola2017image} based discriminator through generative adversarial training strategy and multiple losses including GAN loss, photo loss, perceptual loss, and feature matching loss similar to \cite{liu2023moda}.


\section{Experiments}

\begin{table*}[t]
\centering
\caption{\textbf{Quantitative results of 3D singing head animation.} 
We report the evaluation metrics of the pre-trained models and the models trained on the SingingHead dataset. The units of LVE, FVE, and FDD are $mm$, $mm$, and $\times 10^{-2} mm$, respectively. The metrics of FaceDiffuser are calculated under 10 samples. For the 3D part of the proposed UniSinger framework (Ours), we report the results under 1, 10, and 30 samples, respectively. Note that only methods that generate head pose have the BA and PPE metrics, and only non-deterministic generation methods have the Diversity metric.}
\resizebox{\linewidth}{!}{
\begin{tabular}{c|>{\columncolor[gray]{.95}}l|>{\columncolor[gray]{.95}}c>{\columncolor[gray]{.95}}c>{\columncolor[gray]{.95}}c|>{\columncolor[gray]{.95}}c>{\columncolor[gray]{.95}}c|>{\columncolor[gray]{.95}}c|>{\columncolor[gray]{.95}}c}
\toprule
\rowcolor{white}
& Method  & LVE-Mean/Min\,$\downarrow$  & FVE-Mean/Min\,$\downarrow$  & FDD\,$\downarrow$  & BA\,$\uparrow$  & PPE\,$\downarrow$  & Diversity\,$\uparrow$  & FPS\,$\uparrow$ \\ 
\midrule 
\rowcolor{white}
& FaceFormer~\cite{fan2022faceformer} & 13.025 / 13.025   & 2.8999 / 2.8999 & 2.2496  &-   &-   &-   & 1.6171 \\
\rowcolor{white}
& CodeTalker~\cite{xing2023codetalker} & 13.103 / 13.103   & 2.9212 / 2.9212  & 2.1548   &-   &-   &-   & 0.2209  \\
\rowcolor{white}
& Imitator~\cite{thambiraja2023imitator} & 12.591 / 12.591  & 2.8483 / 2.8483 & 2.9012  &   &   &-   & 1.8329 \\
\rowcolor{white}
\multirow{-4}{*}{Pre-trained} & SelfTalk~\cite{peng2023selftalk} & 12.971 / 12.971  & 2.9018 / 2.9018  & 1.5410    &-   &-   &-   & \underline{28.103} \\
\midrule
\rowcolor{white}
& FaceFormer~\cite{fan2022faceformer} & \underline{4.4715} / 4.4715    & \textbf{1.0255} / 1.0255  & 2.6241  &-  &-   &-   & 1.4500 \\
\rowcolor{white}
& CodeTalker~\cite{xing2023codetalker} & 5.9122 / 5.9122  & 1.3406 / 1.3406 & 2.4946   &-   &-   &-   & 0.2204 \\
\rowcolor{white}
& Imitator~\cite{thambiraja2023imitator} & \textbf{4.4188} / 4.4188  & \underline{1.0265} / 1.0265  & 2.7503   &-   &-   &-   & 1.9076 \\
\rowcolor{white}
& SelfTalk~\cite{peng2023selftalk} & 4.7316 / 4.7316  & 1.0963 / 1.0963  & 1.2911   &-   &-   &-   & 27.360 \\
\rowcolor{white}
& FaceDiffuser~\cite{stan2023facediffuser} & 8.8594 / 7.2877   & 2.1219 / 1.0907  & \textbf{1.1435}  & 0.2152  & 0.2216  & \textbf{3.9516}  & 0.0615 \\
& \textbf{Ours (n=1)}  & 4.5802 / 4.5802  & 1.0412 / 1.0412  & \underline{1.2235}  & \textbf{0.2382}  & \textbf{0.0984}  & -  & \textbf{50.849} \\
& \textbf{Ours (n=10)}  & 4.5928 / 4.2806   & 1.0435 / 0.7909  & 1.2286  & 0.2378  & \underline{0.0988}  & 1.0993  & \textbf{50.849} \\
\multirow{-8}{*}{\shortstack{Trained on\\SingingHead\\Dataset}} & \textbf{Ours (n=30)}  & 4.5958 / \textbf{4.2242}  & 1.0440 / \textbf{0.7232}   & 1.2288  & \underline{0.2381}  & 0.0989  & \underline{1.1006}  & \textbf{50.849} \\
\bottomrule
\end{tabular}
}
\label{tab:3dscores}
\end{table*}

\subsection{Implementation details}
We train the proposed UniSinger framework on our SiningHead dataset. 
Specifically, we cut the sequences in the dataset into equal-length 8s segments, and delete the cropped segments less than 8s in length, which leads to a set of data containing 12196 sequences. Then, the dataset is split into training, validation, and test sets with a ratio of $80\%:5\%:15\%$.
We use all the 9758 sequences in the training set to train the VAE-based 3D facial animation module, and test it on the whole test set which contains 1829 sequences. The training process takes about 2 days on 4 NVIDIA 3090 GPUs.
As for the face renderer, we train a separate network for each person similar to \cite{liu2023moda,ye2023geneface,guo2021ad}. We select 10 identities for evaluation. For each identity, the training of the face renderer takes about 10 hours on a single NVIDIA 3090 GPU.

Moreover, we benchmark existing audio-driven 3D facial animation
methods and 2D talking head methods on the SingingHead dataset. For the 3D singing head animation, we benchmark the pre-trained models \cite{fan2022faceformer,xing2023codetalker,thambiraja2023imitator,peng2023selftalk} using all the 1829 sequences in the SingingHead test set. We also train the existing methods \cite{fan2022faceformer,xing2023codetalker,thambiraja2023imitator,peng2023selftalk,stan2023facediffuser} on the SingingHead training set, and report the results.
For 2D singing portrait video synthesis, we use a total of 267 video sequences from the selected 10 identities as the test set, and benchmark some typical pre-trained 2D talking head models. 
Since our UniSinger framework can handle both 2D and 3D singing head animation tasks, we report its results on 3D singing head animation and 2D singing portrait video synthesis, respectively.

\subsection{3D Singing Head Animation}

\noindent\textbf{Evaluation metrics.} 
We evaluate the 3D singing head animation performance from four aspects. 
(1) \textbf{Facial accuracy.}
Following existing audio-driven 3D facial animation methods, we measure the difference between the generated and the ground truth mesh vertices.
To measure the lip synchronization, we adopt Lip Vertex Error (LVE) \cite{richard2021meshtalk}, which is calculated as the average value of the maximal $L_2$ error of all lip vertices in each frame.
To evaluate the realism of the upper-face dynamics, we calculate the Upper-face Dynamics Deviation (FDD) similar to CodeTalker \cite{xing2023codetalker}.
Moreover, we calculate the average $L_2$ distance of the generated and ground truth mesh vertices in the face region, termed Face Vertex Error (FVE), to measure the accuracy of facial area movements. 
Considering the non-deterministic methods, we report both the mean and minimum values of the LVE and FVE over all the generated samples.
Note that all these metrics are calculated on the mesh vertices without head pose.
(2) \textbf{Head pose.}
We define Pose Parameter Prror (PPE) as the $L_2$ distance between the generated and ground truth 6-dimensional head pose parameters, which measures the accuracy of the head movements.
To measure the alignment between the input singing audio and head movements, we adopt the Beat Align Score (BA) \cite{siyao2022bailando} to the generated head pose parameters.
(3) \textbf{Diversity.}
To measure the diversity of the generation results, we calculate the average $L_2$ distance of 68 facial landmarks between all pairs from the generated $n$ samples, which can be written as $\frac{1}{n(n-1)} {\textstyle \sum_{i=1}^{n}} {\textstyle \sum_{j\ne i}^{n}} ||\boldsymbol{L}_i-\boldsymbol{L}_j||_2^2$.
(4) \textbf{Inference speed.}
It is important to generate 3D facial motion in real time according to the input audio. Therefore, we measure the model inference speed by calculating the average number of frames generated per second, termed FPS.

\noindent\textbf{Baseline methods.} 
We select some typical 3D facial animation methods as the baseline methods, including FaceFormer \cite{fan2022faceformer}, CodeTalker \cite{xing2023codetalker}, Imitator \cite{thambiraja2023imitator}, SelfTalk \cite{peng2023selftalk} and FaceDiffuser \cite{stan2023facediffuser}. We not only evaluate their pre-trained models on the SingingHead test set, but also train them on our SingingHead dataset.
Since most of these methods \cite{fan2022faceformer,xing2023codetalker,thambiraja2023imitator,peng2023selftalk} are designed for mesh sequences, while the 3D facial motion in our dataset is represented by FLAME parameters \cite{li2017learning}, so we use the FLAME mesh sequences without head pose to train these methods.
FaceDiffuser \cite{stan2023facediffuser} and the 3D facial animation module of our UniSinger framework are trained directly using the FLAME parameters which include the head pose parameters.

\begin{figure}
\centering
\includegraphics[width=\linewidth]{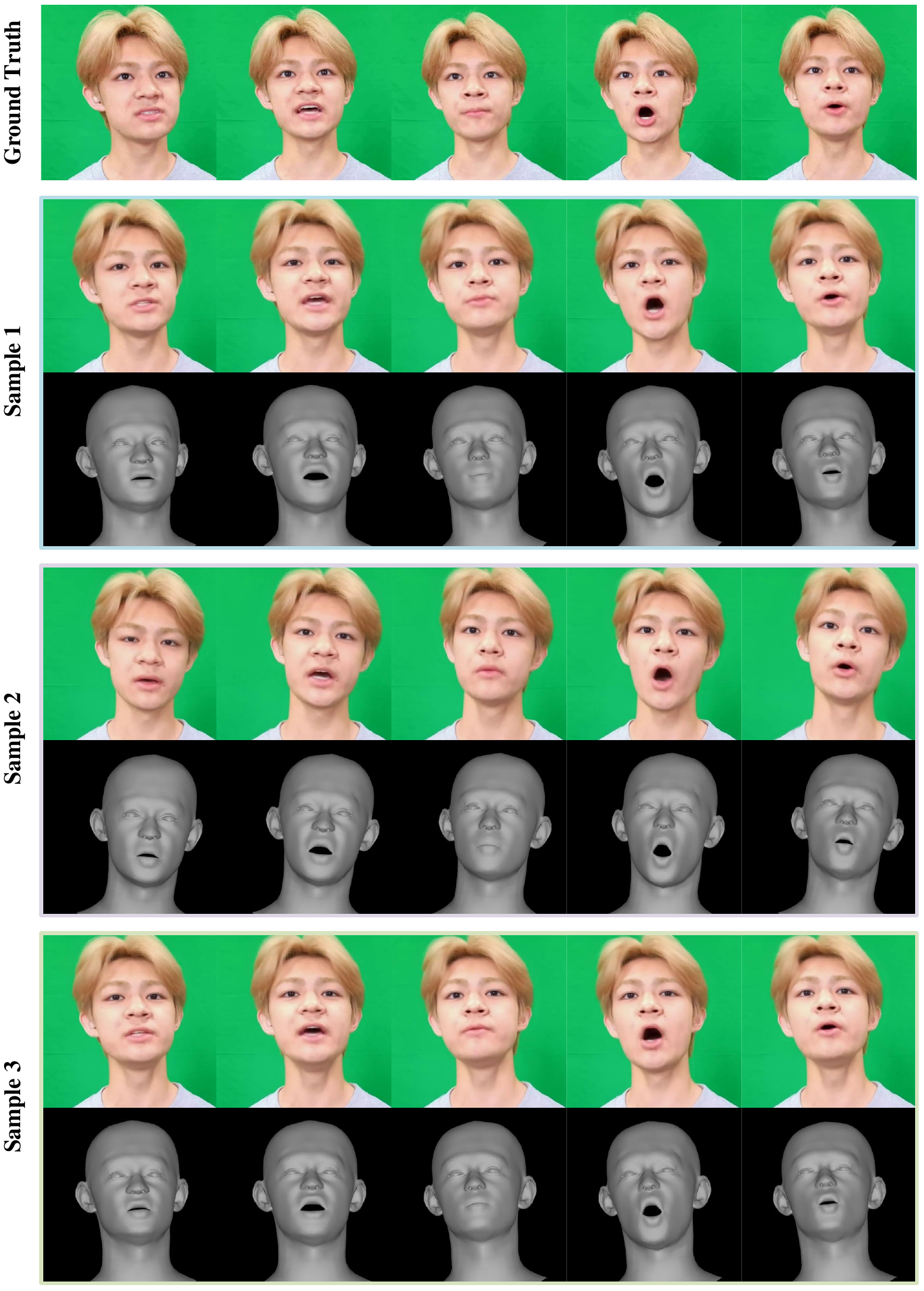}
\caption{\textbf{Illustration of the diverse generation.} Our model can generate diverse 3D singing head motions and corresponding portrait videos from the same input singing audio.}
\label{fig:diversity}
\end{figure}

\begin{figure*}[t]
\centering
\includegraphics[width=\linewidth]{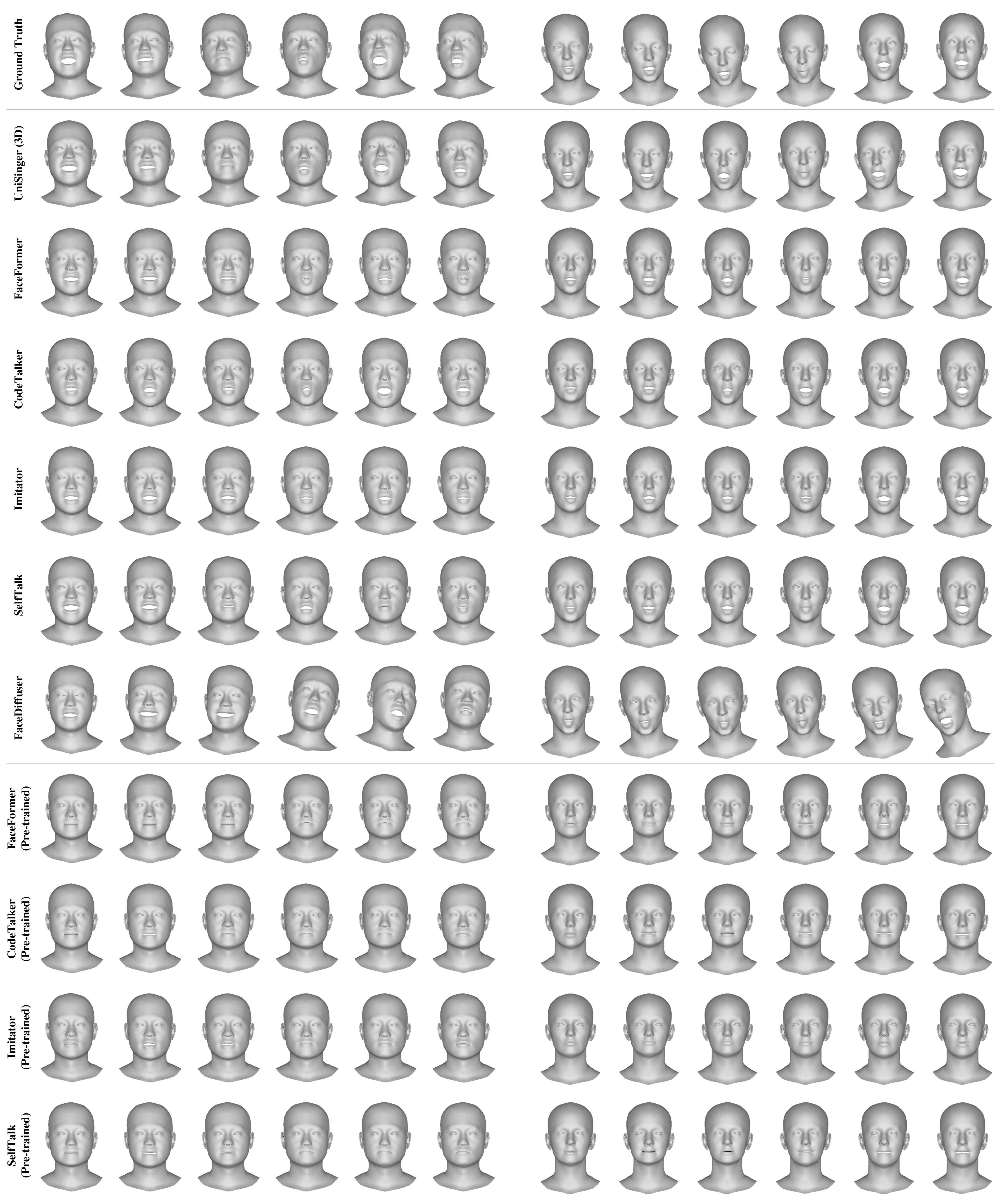}
\caption{\textbf{Qualitative results of 3D singing head animation.} We show the generated 3D facial motion sequences according to two input singing audios. Our method can generate more accurate and natural motion sequences compared to other methods. The pre-trained models perform the worst, which demonstrates the necessity of our singing-specific dataset for the 3D singing head animation task.}
\label{fig:compare3d}
\end{figure*}

The quantitative and qualitative results are shown in Tab. \ref{tab:3dscores} and Fig. \ref{fig:compare3d}, respectively. 
On the one hand, pre-trained models show the worst quantitative and qualitative results, which indicates that the 3D facial animation models trained on talking datasets cannot generalize well to singing audio. After training on the SingingHead dataset, almost all the metrics and visualization results of these models have improved, which demonstrates the necessity of our singing-specific dataset for the 3D singing head animation task.
On the other hand, looking at all the models that are trained on the SingingHead dataset, our method achieves more realistic and vivid visual results, and competitive results on all metrics. 
Specifically, nearly all the previous methods \cite{fan2022faceformer,xing2023codetalker,thambiraja2023imitator,peng2023selftalk} use offsets of mesh vertices to represent the facial movements, which is not only computationally intensive but also hard to generate head movements. In contrast, our method uses sequences of the 59-dimensional FLAME parameters \cite{li2017learning} including head pose parameters to represent motions, leading to more accurate head movements and vivid results as evidenced by the PPE and BA metrics.
Moreover, most of the existing methods \cite{fan2022faceformer,xing2023codetalker,thambiraja2023imitator,peng2023selftalk} model the audio-driven 3D facial animation as a deterministic regression task, which can only generate one results given input audio. Such methods can regress more accurate vertex positions, leading to better vertex-level metrics such as LVE and FVE. In contrast, FaceDiffuser \cite{stan2023facediffuser} and our method model it as a generation task, which can generate diverse facial motions according to one input audio. Generative models tend to produce inferior results on deterministic metrics. Nevertheless, our method achieves competitive results on LVE and FVE metrics and demonstrates some diversity in generation as shown in Fig. \ref{fig:diversity}. Achieving accuracy and diversity in generation involves a trade-off. Although our method exhibits less diversity compared to FaceDiffuser \cite{stan2023facediffuser}, it achieves much better lip and face synchronization.
In addition, our method has the fastest inference speed and can achieve real-time 3D singing head animation, thanks to the non-autoregressive mechanism we employed.

We also conduct ablation studies on the proposed 3D facial animation module of the UniSinger framework as shown in Tab. \ref{tab:ablation}. 
The injection of background music helps the model generate more accurate facial and head movements, while slightly reducing the variety of generated results due to the additional conditions introduced.
Reconstruction loss $\mathcal{L}_{rc}$ and KL loss $\mathcal{L}_{kl}$ are necessary for the training of VAE. Without $\mathcal{L}_{rc}$, the latent space of the model tends to concentrate in a small region of the prior distribution, resulting in limited diverse generation ability. Without $\mathcal{L}_{kl}$, the model is hard to converge, leading to poor accuracy and diversity in generation.
Velocity loss $\mathcal{L}_{vel}$ is used to prevent the generation of static or motionless results. 
Mesh vertices loss $\mathcal{L}_{mesh}$ contributes to increasing the realism and diversity of the generated results, as well as improving the alignment with the input audio.

\begin{table}[h]
\centering
\caption{\textbf{Ablation study for vae-based 3D facial animation module.} The units of LVE, FVE, and FDD are $mm$, $mm$, and $\times 10^{-2} mm$, respectively.}
\resizebox{\linewidth}{!}{
\begin{tabular}{l|cccccc}
\toprule
Method  & LVE\,$\downarrow$  & FVE\,$\downarrow$  & FDD\,$\downarrow$  & BA\,$\uparrow$  & PPE\,$\downarrow$  & Diversity\,$\uparrow$ \\ 
\midrule 
w/o BGM & 4.8025  & 1.1017  & 1.1874  & 0.2343  & 0.1042  & 1.2407  \\
w/o $L_{rc}$ & 4.4723   & 1.0035  & 1.4530   & 0.2288   & 0.1475   & 0.0042  \\
w/o $L_{kl}$ & 4.9272   & 1.1274  & 0.8862   & 0.2402   & 0.1031  & 0.5593  \\
w/o $L_{vel}$ & 4.5952  & 1.0397  & 1.1991  & 0.2389   & 0.0966  & 0.9608 \\
w/o $L_{mesh}$ & 4.5611  & 1.0230  & 1.2532  & 0.2355   & 0.0933  & 0.7873 \\
\rowcolor[gray]{.95}
Ours & 4.5928  & 1.0435  & 1.2286  & 0.2378  & 0.0988  & 1.0993 \\
\bottomrule
\end{tabular}
}
\label{tab:ablation}
\end{table}

\begin{table*}
\centering
\caption{\textbf{Quantitative results of 2D singing portrait video synthesis.} We report the evaluation metrics on the generated singing videos of the selected 10 identities. We list results under the resolution of 256 / 512, respectively.}
\resizebox{\linewidth}{!}{
\begin{tabular}{l|cccc|cc|c}
\toprule
Method  & FID\,$\downarrow$  & LPIPS\,$\downarrow$  & SSIM\,$\uparrow$  & CPBD\,$\uparrow$  & M-LMD\,$\downarrow$  & F-LMD\,$\downarrow$  & BA\,$\uparrow$ \\ 
\midrule 
Ground Truth & 0.000  & 0.000  & 1.000  & 0.401 / 0.299  & 0.000   & 0.000   & 0.252 / 0.249 \\
MakeItTalk~\cite{zhou2020makelttalk} & 29.21 / 46.90    & 0.180 / 0.276  & \textbf{0.620} / \textbf{0.690}  & 0.226 / 0.034  & 2.753 / 5.419   & \underline{4.510} / \underline{8.972}   & 0.243 / 0.238 \\
Audio2Head~\cite{wang2021audio2head} & 42.06 / 62.98  & 0.233 / 0.328  & 0.560 / 0.647   & 0.228 / 0.044   & 2.711 / 5.343   & 5.002 / 10.01   & \underline{0.249} / \textbf{0.254} \\
SadTalker~\cite{zhang2023sadtalker} & \underline{18.07} / \underline{18.73}  &  \underline{0.175} / \underline{0.238}  & 0.592 / 0.660   & 0.217 / \underline{0.161}   & 2.989 / 5.887   & 4.774 / 9.508   & 0.237 / 0.239 \\
EAT~\cite{gan2023efficient} & 54.28 / 75.58  & 0.238 / 0.340  & 0.553 / 0.640   & 0.230 / 0.048   & \underline{2.672} / \underline{5.248}   & 5.112 / 10.18   & 0.244 / 0.243 \\
DreamTalk~\cite{ma2023dreamtalk} & 53.91 / 66.25  & 0.297 / 0.404  & 0.492 / 0.592  & \underline{0.245} / 0.048  & 3.655 / 7.221  & 8.261 / 16.42  & 0.221 / 0.219 \\
\rowcolor[gray]{.95}
\textbf{UniSinger (2D)}  & \textbf{12.31} / \textbf{12.08}  & \textbf{0.169} / \textbf{0.233}  & \underline{0.618} / \underline{0.688}  & \textbf{0.258} / \textbf{0.165}  & \textbf{2.041} / \textbf{3.991}  & \textbf{3.527} / \textbf{7.040}  & \textbf{0.251} / \underline{0.251} \\
\bottomrule
\end{tabular}
}
\label{tab:2dscore}
\vspace{-2mm}
\end{table*}

\begin{figure*}[h]
\centering
\includegraphics[width=\linewidth]{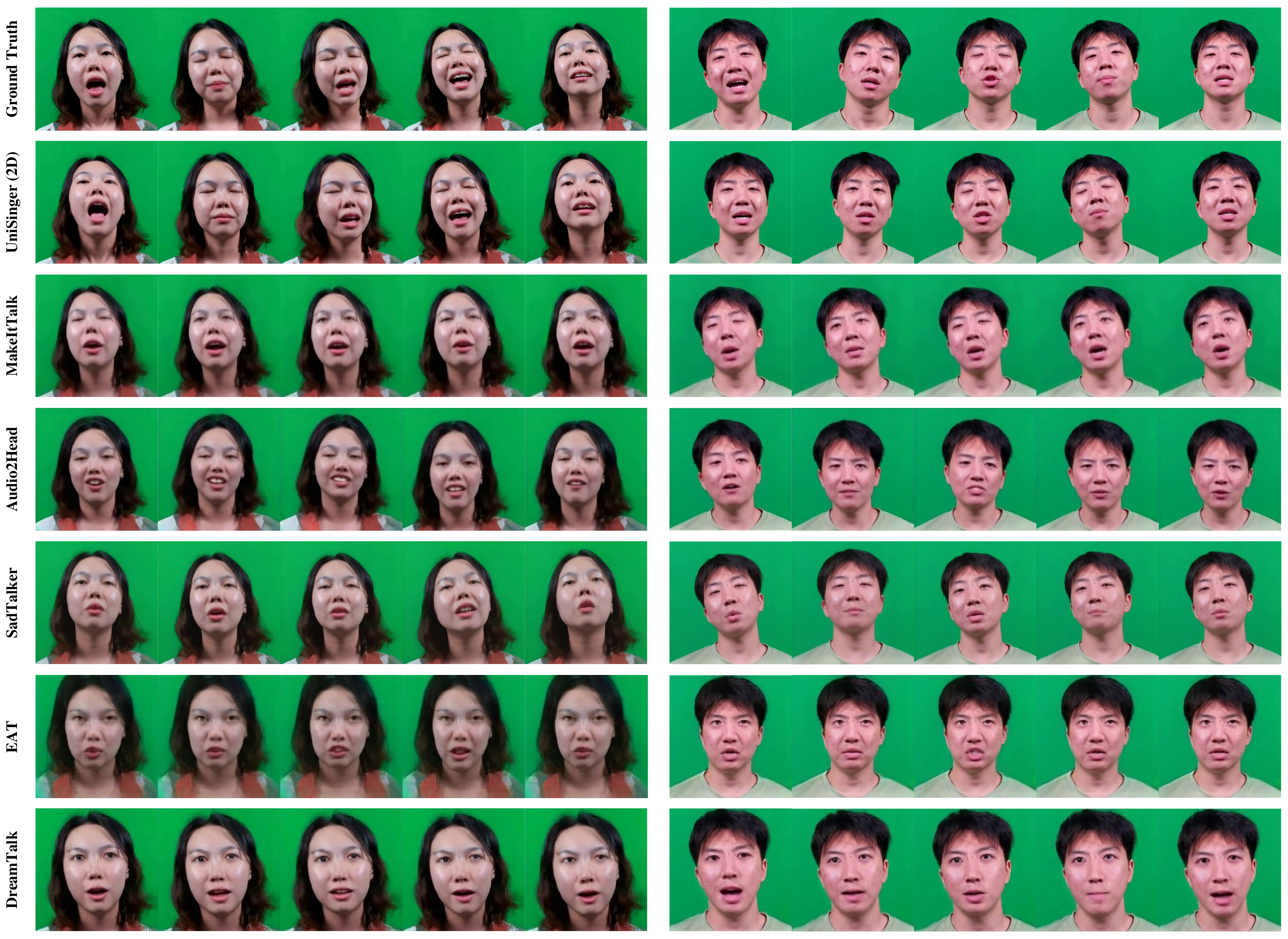}
\caption{\textbf{Qualitative results of 2D singing portrait video synthesis.} Our method can generate singing portrait videos with higher quality and accuracy in terms of facial movements, especially in the mouth region, identity preservation, and video quality.}
\label{fig:compare2d}
\vspace{-1mm}
\end{figure*}





\subsection{2D Singing Portrait Video Synthesis}

\noindent\textbf{Evaluation metrics.}
We evaluate the performance of 2D singing portrait video synthesis from three aspects.
(1) \textbf{Video quality.}
We use full-reference metrics, including Frechet Inception Distance (FID) \cite{heusel2017gans,parmar2021cleanfid}, Learned Perceptual Image Patch Similarity (LPIPS) \cite{zhang2018unreasonable} and Structural Similarity (SSIM) \cite{wang2004image}, to measure the realism, perceptual and structural similarity of the generated video frames compared to the ground truth.
Besides, the no-reference metric Cumulative Probability Blur Detection (CPBD) \cite{narvekar2011no} is adopted to evaluate the sharpness of the generated frames.
(2) \textbf{Accuracy of facial movements.}
We adopt Landmark Distance (LMD) \cite{chen2018lip} to evaluate the alignment of generated facial movements with the ground truth, thereby assessing the synchronization of the generated portrait video with the input audio.
Specifically, we first detect the 68 facial landmarks in both generated and ground truth frames using \cite{bulat2017far}. Then, we calculate the distance between the generated and ground truth landmarks in the mouth region and the entire face region to obtain the metrics M-LMD and F-LMD, respectively. 
(3) \textbf{Alignment with audio.}
To measure whether the facial movements in the generated video align well with the input audio, we adopt the Beat Align Score (BA) \cite{siyao2022bailando} that is applied to the 68 facial landmarks.

\noindent\textbf{Baseline methods.}
We select some classic 2D talking head methods including MakeItTalk \cite{zhou2020makelttalk}, Audio2Head \cite{wang2021audio2head}, SadTalker \cite{zhang2023sadtalker}, EAT \cite{gan2023efficient} and DreamTalk \cite{ma2023dreamtalk}.
Since the 2D talking head methods are typically trained on large-scale datasets and have zero-shot capability, we directly evaluate their pre-trained models on the selected identities of the SingingHead test set. 

\clearpage
The quantitative and qualitative results are shown in Tab. \ref{tab:2dscore} and Fig. \ref{fig:compare2d}, respectively. 
We can see that our method achieves much more accurate facial movements and better alignment with input audio. Although other pre-trained models are trained on very large-scale talking head datasets, they often fail to generalize well to the singing task. As a result, these methods struggle with lip synchronization, especially in scenarios where the mouth needs to be opened wider, which is not common in regular talking.
Therefore, it is essential to use singing-specific data to help existing 2D talking head methods generalize well to the singing task and achieve better performance in 2D singing portrait video synthesis.
Moreover, our method also achieves higher video quality and more realistic and vivid visualization results.

\section{Conclusion}
In this paper, we focus on the task of singing head animation and collect the first large-scale high-quality 4D facial animation dataset specialized for singing, named \textbf{SingingHead}. This dataset contains more than 27 hours' synchronized portrait video, 3D facial motion sequence, singing audio, and background music from 76 subjects.
Along with the dataset, we benchmark existing audio-driven 3D facial animation methods and 2D talking head methods on the singing task.
Moreover, we propose a unified framework named \textbf{UniSinger} to achieve both singing
audio-driven 3D singing head animation and 2D singing portrait video synthesis, which achieves competitive results on both 3D and 2D benchmarks.

\bibliographystyle{IEEEtran}
\bibliography{refs}

\begin{thebibliography}{10}
\providecommand{\url}[1]{#1}
\csname url@samestyle\endcsname
\providecommand{\newblock}{\relax}
\providecommand{\bibinfo}[2]{#2}
\providecommand{\BIBentrySTDinterwordspacing}{\spaceskip=0pt\relax}
\providecommand{\BIBentryALTinterwordstretchfactor}{4}
\providecommand{\BIBentryALTinterwordspacing}{\spaceskip=\fontdimen2\font plus
\BIBentryALTinterwordstretchfactor\fontdimen3\font minus \fontdimen4\font\relax}
\providecommand{\BIBforeignlanguage}[2]{{%
\expandafter\ifx\csname l@#1\endcsname\relax
\typeout{** WARNING: IEEEtran.bst: No hyphenation pattern has been}%
\typeout{** loaded for the language `#1'. Using the pattern for}%
\typeout{** the default language instead.}%
\else
\language=\csname l@#1\endcsname
\fi
#2}}
\providecommand{\BIBdecl}{\relax}
\BIBdecl

\bibitem{zhang2023sadtalker}
W.~Zhang, X.~Cun, X.~Wang, Y.~Zhang, X.~Shen, Y.~Guo, Y.~Shan, and F.~Wang, ``Sadtalker: Learning realistic 3d motion coefficients for stylized audio-driven single image talking face animation,'' in \emph{Proceedings of the IEEE/CVF Conference on Computer Vision and Pattern Recognition}, 2023, pp. 8652--8661.

\bibitem{liu2023moda}
Y.~Liu, L.~Lin, F.~Yu, C.~Zhou, and Y.~Li, ``Moda: Mapping-once audio-driven portrait animation with dual attentions,'' in \emph{Proceedings of the IEEE/CVF International Conference on Computer Vision}, 2023, pp. 23\,020--23\,029.

\bibitem{liu2024audio}
M.~Liu, D.~Li, Y.~Li, X.~Song, and L.~Nie, ``Audio-semantic enhanced pose-driven talking head generation,'' \emph{IEEE Transactions on Circuits and Systems for Video Technology}, 2024.

\bibitem{fan2022faceformer}
Y.~Fan, Z.~Lin, J.~Saito, W.~Wang, and T.~Komura, ``Faceformer: Speech-driven 3d facial animation with transformers,'' in \emph{Proceedings of the IEEE/CVF Conference on Computer Vision and Pattern Recognition}, 2022, pp. 18\,770--18\,780.

\bibitem{xing2023codetalker}
J.~Xing, M.~Xia, Y.~Zhang, X.~Cun, J.~Wang, and T.-T. Wong, ``Codetalker: Speech-driven 3d facial animation with discrete motion prior,'' in \emph{Proceedings of the IEEE/CVF Conference on Computer Vision and Pattern Recognition}, 2023, pp. 12\,780--12\,790.

\bibitem{chu2024corrtalk}
Z.~Chu, K.~Guo, X.~Xing, Y.~Lan, B.~Cai, and X.~Xu, ``Corrtalk: Correlation between hierarchical speech and facial activity variances for 3d animation,'' \emph{IEEE Transactions on Circuits and Systems for Video Technology}, 2024.

\bibitem{wu2023mmface4d}
H.~Wu, J.~Jia, J.~Xing, H.~Xu, X.~Wang, and J.~Wang, ``Mmface4d: A large-scale multi-modal 4d face dataset for audio-driven 3d face animation,'' \emph{arXiv preprint arXiv:2303.09797}, 2023.

\bibitem{zhang2021flow}
Z.~Zhang, L.~Li, Y.~Ding, and C.~Fan, ``Flow-guided one-shot talking face generation with a high-resolution audio-visual dataset,'' in \emph{Proceedings of the IEEE/CVF Conference on Computer Vision and Pattern Recognition}, 2021, pp. 3661--3670.

\bibitem{wang2020mead}
K.~Wang, Q.~Wu, L.~Song, Z.~Yang, W.~Wu, C.~Qian, R.~He, Y.~Qiao, and C.~C. Loy, ``Mead: A large-scale audio-visual dataset for emotional talking-face generation,'' in \emph{European Conference on Computer Vision}.\hskip 1em plus 0.5em minus 0.4em\relax Springer, 2020, pp. 700--717.

\bibitem{fanelli20103}
G.~Fanelli, J.~Gall, H.~Romsdorfer, T.~Weise, and L.~Van~Gool, ``A 3-d audio-visual corpus of affective communication,'' \emph{IEEE Transactions on Multimedia}, vol.~12, no.~6, pp. 591--598, 2010.

\bibitem{cudeiro2019capture}
D.~Cudeiro, T.~Bolkart, C.~Laidlaw, A.~Ranjan, and M.~J. Black, ``Capture, learning, and synthesis of 3d speaking styles,'' in \emph{Proceedings of the IEEE/CVF Conference on Computer Vision and Pattern Recognition}, 2019, pp. 10\,101--10\,111.

\bibitem{thambiraja2023imitator}
B.~Thambiraja, I.~Habibie, S.~Aliakbarian, D.~Cosker, C.~Theobalt, and J.~Thies, ``Imitator: Personalized speech-driven 3d facial animation,'' in \emph{Proceedings of the IEEE/CVF International Conference on Computer Vision}, 2023, pp. 20\,621--20\,631.

\bibitem{peng2023selftalk}
Z.~Peng, Y.~Luo, Y.~Shi, H.~Xu, X.~Zhu, H.~Liu, J.~He, and Z.~Fan, ``Selftalk: A self-supervised commutative training diagram to comprehend 3d talking faces,'' in \emph{Proceedings of the 31st ACM International Conference on Multimedia}, 2023, pp. 5292--5301.

\bibitem{stan2023facediffuser}
S.~Stan, K.~I. Haque, and Z.~Yumak, ``Facediffuser: Speech-driven 3d facial animation synthesis using diffusion,'' in \emph{Proceedings of the 16th ACM SIGGRAPH Conference on Motion, Interaction and Games}, 2023, pp. 1--11.

\bibitem{wang2021audio2head}
S.~Wang, L.~Li, Y.~Ding, C.~Fan, and X.~Yu, ``Audio2head: Audio-driven one-shot talking-head generation with natural head motion,'' \emph{arXiv preprint arXiv:2107.09293}, 2021.

\bibitem{gan2023efficient}
Y.~Gan, Z.~Yang, X.~Yue, L.~Sun, and Y.~Yang, ``Efficient emotional adaptation for audio-driven talking-head generation,'' in \emph{Proceedings of the IEEE/CVF International Conference on Computer Vision}, 2023, pp. 22\,634--22\,645.

\bibitem{ma2023dreamtalk}
Y.~Ma, S.~Zhang, J.~Wang, X.~Wang, Y.~Zhang, and Z.~Deng, ``Dreamtalk: When expressive talking head generation meets diffusion probabilistic models,'' \emph{arXiv preprint arXiv:2312.09767}, 2023.

\bibitem{richard2021meshtalk}
A.~Richard, M.~Zollh{\"o}fer, Y.~Wen, F.~De~la Torre, and Y.~Sheikh, ``Meshtalk: 3d face animation from speech using cross-modality disentanglement,'' in \emph{Proceedings of the IEEE/CVF International Conference on Computer Vision}, 2021, pp. 1173--1182.

\bibitem{peng2023emotalk}
Z.~Peng, H.~Wu, Z.~Song, H.~Xu, X.~Zhu, J.~He, H.~Liu, and Z.~Fan, ``Emotalk: Speech-driven emotional disentanglement for 3d face animation,'' in \emph{Proceedings of the IEEE/CVF International Conference on Computer Vision}, 2023, pp. 20\,687--20\,697.

\bibitem{danvevcek2023emotional}
R.~Dan{\v{e}}{\v{c}}ek, K.~Chhatre, S.~Tripathi, Y.~Wen, M.~Black, and T.~Bolkart, ``Emotional speech-driven animation with content-emotion disentanglement,'' in \emph{SIGGRAPH Asia 2023 Conference Papers}, 2023, pp. 1--13.

\bibitem{sun2023diffposetalk}
Z.~Sun, T.~Lv, S.~Ye, M.~G. Lin, J.~Sheng, Y.-H. Wen, M.~Yu, and Y.-j. Liu, ``Diffposetalk: Speech-driven stylistic 3d facial animation and head pose generation via diffusion models,'' \emph{arXiv preprint arXiv:2310.00434}, 2023.

\bibitem{zhao2024media2face}
Q.~Zhao, P.~Long, Q.~Zhang, D.~Qin, H.~Liang, L.~Zhang, Y.~Zhang, J.~Yu, and L.~Xu, ``Media2face: Co-speech facial animation generation with multi-modality guidance,'' \emph{arXiv preprint arXiv:2401.15687}, 2024.

\bibitem{yu2020multimodal}
L.~Yu, J.~Yu, M.~Li, and Q.~Ling, ``Multimodal inputs driven talking face generation with spatial--temporal dependency,'' \emph{IEEE Transactions on Circuits and Systems for Video Technology}, vol.~31, no.~1, pp. 203--216, 2020.

\bibitem{prajwal2020lip}
K.~Prajwal, R.~Mukhopadhyay, V.~P. Namboodiri, and C.~Jawahar, ``A lip sync expert is all you need for speech to lip generation in the wild,'' in \emph{Proceedings of the 28th ACM international conference on multimedia}, 2020, pp. 484--492.

\bibitem{zhou2020makelttalk}
Y.~Zhou, X.~Han, E.~Shechtman, J.~Echevarria, E.~Kalogerakis, and D.~Li, ``Makelttalk: speaker-aware talking-head animation,'' \emph{ACM Transactions On Graphics (TOG)}, vol.~39, no.~6, pp. 1--15, 2020.

\bibitem{zhou2021pose}
H.~Zhou, Y.~Sun, W.~Wu, C.~C. Loy, X.~Wang, and Z.~Liu, ``Pose-controllable talking face generation by implicitly modularized audio-visual representation,'' in \emph{Proceedings of the IEEE/CVF conference on computer vision and pattern recognition}, 2021, pp. 4176--4186.

\bibitem{ye2023geneface}
Z.~Ye, Z.~Jiang, Y.~Ren, J.~Liu, J.~He, and Z.~Zhao, ``Geneface: Generalized and high-fidelity audio-driven 3d talking face synthesis,'' \emph{arXiv preprint arXiv:2301.13430}, 2023.

\bibitem{shen2023difftalk}
S.~Shen, W.~Zhao, Z.~Meng, W.~Li, Z.~Zhu, J.~Zhou, and J.~Lu, ``Difftalk: Crafting diffusion models for generalized audio-driven portraits animation,'' in \emph{Proceedings of the IEEE/CVF Conference on Computer Vision and Pattern Recognition}, 2023, pp. 1982--1991.

\bibitem{sheng2023stochastic}
Z.~Sheng, L.~Nie, M.~Zhang, X.~Chang, and Y.~Yan, ``Stochastic latent talking face generation towards emotional expressions and head poses,'' \emph{IEEE Transactions on Circuits and Systems for Video Technology}, 2023.

\bibitem{liu2024osm}
J.~Liu, X.~Wang, X.~Fu, Y.~Chai, C.~Yu, J.~Dai, and J.~Han, ``Osm-net: One-to-many one-shot talking head generation with spontaneous head motions,'' \emph{IEEE Transactions on Circuits and Systems for Video Technology}, 2024.

\bibitem{zhang2024hierarchical}
J.~Zhang, C.~Liu, K.~Xian, and Z.~Cao, ``Hierarchical feature warping and blending for talking head animation,'' \emph{IEEE Transactions on Circuits and Systems for Video Technology}, 2024.

\bibitem{iwase2020song2face}
S.~Iwase, T.~Kato, S.~Yamaguchi, T.~Yukitaka, and S.~Morishima, ``Song2face: Synthesizing singing facial animation from audio,'' in \emph{SIGGRAPH Asia 2020 Technical Communications}, 2020, pp. 1--4.

\bibitem{liu2023musicface}
P.~Liu, W.~Deng, H.~Li, J.~Wang, Y.~Zheng, Y.~Ding, X.~Guo, and M.~Zeng, ``Musicface: Music-driven expressive singing face synthesis,'' \emph{arXiv preprint arXiv:2303.14044}, 2023.

\bibitem{livingstone2018ryerson}
S.~R. Livingstone and F.~A. Russo, ``The ryerson audio-visual database of emotional speech and song (ravdess): A dynamic, multimodal set of facial and vocal expressions in north american english,'' \emph{PloS one}, vol.~13, no.~5, p. e0196391, 2018.

\bibitem{li2017learning}
T.~Li, T.~Bolkart, M.~J. Black, H.~Li, and J.~Romero, ``Learning a model of facial shape and expression from 4d scans.'' \emph{ACM Trans. Graph.}, vol.~36, no.~6, pp. 194--1, 2017.

\bibitem{vaswani2017attention}
A.~Vaswani, N.~Shazeer, N.~Parmar, J.~Uszkoreit, L.~Jones, A.~N. Gomez, {\L}.~Kaiser, and I.~Polosukhin, ``Attention is all you need,'' \emph{Advances in neural information processing systems}, vol.~30, 2017.

\bibitem{kingma2013auto}
D.~P. Kingma and M.~Welling, ``Auto-encoding variational bayes,'' \emph{arXiv preprint arXiv:1312.6114}, 2013.

\bibitem{petrovich2021action}
M.~Petrovich, M.~J. Black, and G.~Varol, ``Action-conditioned 3d human motion synthesis with transformer vae,'' in \emph{Proceedings of the IEEE/CVF International Conference on Computer Vision}, 2021, pp. 10\,985--10\,995.

\bibitem{cao2005expressive}
Y.~Cao, W.~C. Tien, P.~Faloutsos, and F.~Pighin, ``Expressive speech-driven facial animation,'' \emph{ACM Transactions on Graphics (TOG)}, vol.~24, no.~4, pp. 1283--1302, 2005.

\bibitem{taylor2017deep}
S.~Taylor, T.~Kim, Y.~Yue, M.~Mahler, J.~Krahe, A.~G. Rodriguez, J.~Hodgins, and I.~Matthews, ``A deep learning approach for generalized speech animation,'' \emph{ACM Transactions On Graphics (TOG)}, vol.~36, no.~4, pp. 1--11, 2017.

\bibitem{karras2017audio}
T.~Karras, T.~Aila, S.~Laine, A.~Herva, and J.~Lehtinen, ``Audio-driven facial animation by joint end-to-end learning of pose and emotion,'' \emph{ACM Transactions on Graphics (TOG)}, vol.~36, no.~4, pp. 1--12, 2017.

\bibitem{zhou2018visemenet}
Y.~Zhou, Z.~Xu, C.~Landreth, E.~Kalogerakis, S.~Maji, and K.~Singh, ``Visemenet: Audio-driven animator-centric speech animation,'' \emph{ACM Transactions on Graphics (TOG)}, vol.~37, no.~4, pp. 1--10, 2018.

\bibitem{wu2023ganhead}
S.~Wu, Y.~Yan, Y.~Li, Y.~Cheng, W.~Zhu, K.~Gao, X.~Li, and G.~Zhai, ``Ganhead: Towards generative animatable neural head avatars,'' in \emph{Proceedings of the IEEE/CVF conference on computer vision and pattern recognition}, 2023, pp. 437--447.

\bibitem{xiang2023flashavatar}
J.~Xiang, X.~Gao, Y.~Guo, and J.~Zhang, ``Flashavatar: High-fidelity digital avatar rendering at 300fps,'' \emph{arXiv preprint arXiv:2312.02214}, 2023.

\bibitem{chen2023monogaussianavatar}
Y.~Chen, L.~Wang, Q.~Li, H.~Xiao, S.~Zhang, H.~Yao, and Y.~Liu, ``Monogaussianavatar: Monocular gaussian point-based head avatar,'' in \emph{ACM SIGGRAPH Conference Proceedings}, 2024.

\bibitem{zhu2024dfie3d}
X.~Zhu, J.~Zhou, L.~You, X.~Yang, J.~Chang, J.~J. Zhang, and D.~Zeng, ``Dfie3d: 3d-aware disentangled face inversion and editing via facial-contrastive learning,'' \emph{IEEE Transactions on Circuits and Systems for Video Technology}, 2024.

\bibitem{jun2016real}
Y.~Jun, C.~Jiang, R.~Li, C.-W. Luo, and Z.-F. Wang, ``Real-time 3-d facial animation: From appearance to internal articulators,'' \emph{IEEE Transactions on Circuits and Systems for Video Technology}, vol.~28, no.~4, pp. 920--932, 2016.

\bibitem{zhong2024expclip}
Y.~Zhong, H.~Wei, P.~Yang, and Z.~Wang, ``Expclip: Bridging text and facial expressions via semantic alignment,'' in \emph{Proceedings of the AAAI Conference on Artificial Intelligence}, vol.~38, no.~7, 2024, pp. 7614--7622.

\bibitem{aneja2023facetalk}
S.~Aneja, J.~Thies, A.~Dai, and M.~Nie{\ss}ner, ``Facetalk: Audio-driven motion diffusion for neural parametric head models,'' \emph{arXiv preprint arXiv:2312.08459}, 2023.

\bibitem{van2017neural}
A.~Van Den~Oord, O.~Vinyals \emph{et~al.}, ``Neural discrete representation learning,'' \emph{Advances in neural information processing systems}, vol.~30, 2017.

\bibitem{ho2020denoising}
J.~Ho, A.~Jain, and P.~Abbeel, ``Denoising diffusion probabilistic models,'' \emph{Advances in neural information processing systems}, vol.~33, pp. 6840--6851, 2020.

\bibitem{wu2023singinghead}
S.~Wu, Y.~Li, W.~Zhang, J.~Jia, Y.~Zhu, Y.~Yan, and G.~Zhai, ``Singinghead: A large-scale 4d dataset for singing head animation,'' \emph{arXiv preprint arXiv:2312.04369v1}, 2023.

\bibitem{zhou2012image}
Z.~Zhou, G.~Zhao, Y.~Guo, and M.~Pietikainen, ``An image-based visual speech animation system,'' \emph{IEEE Transactions on Circuits and Systems for Video Technology}, vol.~22, no.~10, pp. 1420--1432, 2012.

\bibitem{son2017lip}
J.~Son~Chung, A.~Senior, O.~Vinyals, and A.~Zisserman, ``Lip reading sentences in the wild,'' in \emph{Proceedings of the IEEE conference on computer vision and pattern recognition}, 2017, pp. 6447--6456.

\bibitem{edwards2016jali}
P.~Edwards, C.~Landreth, E.~Fiume, and K.~Singh, ``Jali: an animator-centric viseme model for expressive lip synchronization,'' \emph{ACM Transactions on graphics (TOG)}, vol.~35, no.~4, pp. 1--11, 2016.

\bibitem{zhu2018arbitrary}
H.~Zhu, H.~Huang, Y.~Li, A.~Zheng, and R.~He, ``Arbitrary talking face generation via attentional audio-visual coherence learning,'' \emph{arXiv preprint arXiv:1812.06589}, 2018.

\bibitem{song2022audio}
L.~Song, W.~Wu, C.~Fu, C.~C. Loy, and R.~He, ``Audio-driven dubbing for user generated contents via style-aware semi-parametric synthesis,'' \emph{IEEE Transactions on Circuits and Systems for Video Technology}, vol.~33, no.~3, pp. 1247--1261, 2022.

\bibitem{zhang2023metaportrait}
B.~Zhang, C.~Qi, P.~Zhang, B.~Zhang, H.~Wu, D.~Chen, Q.~Chen, Y.~Wang, and F.~Wen, ``Metaportrait: Identity-preserving talking head generation with fast personalized adaptation,'' in \emph{Proceedings of the IEEE/CVF Conference on Computer Vision and Pattern Recognition}, 2023, pp. 22\,096--22\,105.

\bibitem{peng2024synctalk}
Z.~Peng, W.~Hu, Y.~Shi, X.~Zhu, X.~Zhang, H.~Zhao, J.~He, H.~Liu, and Z.~Fan, ``Synctalk: The devil is in the synchronization for talking head synthesis,'' in \emph{Proceedings of the IEEE/CVF Conference on Computer Vision and Pattern Recognition}, 2024, pp. 666--676.

\bibitem{tian2024emo}
L.~Tian, Q.~Wang, B.~Zhang, and L.~Bo, ``Emo: Emote portrait alive-generating expressive portrait videos with audio2video diffusion model under weak conditions,'' \emph{arXiv preprint arXiv:2402.17485}, 2024.

\bibitem{xu2024vasa}
S.~Xu, G.~Chen, Y.-X. Guo, J.~Yang, C.~Li, Z.~Zang, Y.~Zhang, X.~Tong, and B.~Guo, ``Vasa-1: Lifelike audio-driven talking faces generated in real time,'' \emph{arXiv preprint arXiv:2404.10667}, 2024.

\bibitem{xu2024hallo}
M.~Xu, H.~Li, Q.~Su, H.~Shang, L.~Zhang, C.~Liu, J.~Wang, L.~Van~Gool, Y.~Yao, and S.~Zhu, ``Hallo: Hierarchical audio-driven visual synthesis for portrait image animation,'' \emph{arXiv preprint arXiv:2406.08801}, 2024.

\bibitem{Chung2016lip_sync}
J.~S. Chung and A.~Zisserman, ``Out of time: Automated lip sync in the wild,'' in \emph{Computer Vision - {ACCV} 2016 Workshops - {ACCV} 2016 International Workshops, Taipei, Taiwan, November 20-24, 2016, Revised Selected Papers, Part {II}}, 2016, pp. 251--263.

\bibitem{guo2021ad}
Y.~Guo, K.~Chen, S.~Liang, Y.-J. Liu, H.~Bao, and J.~Zhang, ``Ad-nerf: Audio driven neural radiance fields for talking head synthesis,'' in \emph{Proceedings of the IEEE/CVF International Conference on Computer Vision}, 2021, pp. 5784--5794.

\bibitem{yao2022dfa}
S.~Yao, R.~Zhong, Y.~Yan, G.~Zhai, and X.~Yang, ``Dfa-nerf: Personalized talking head generation via disentangled face attributes neural rendering,'' \emph{arXiv preprint arXiv:2201.00791}, 2022.

\bibitem{liu2022semantic}
X.~Liu, Y.~Xu, Q.~Wu, H.~Zhou, W.~Wu, and B.~Zhou, ``Semantic-aware implicit neural audio-driven video portrait generation,'' in \emph{European Conference on Computer Vision}.\hskip 1em plus 0.5em minus 0.4em\relax Springer, 2022, pp. 106--125.

\bibitem{stypulkowski2023diffused}
M.~Stypulkowski, K.~Vougioukas, S.~He, M.~Zieba, S.~Petridis, and M.~Pantic, ``Diffused heads: Diffusion models beat gans on talking-face generation,'' \emph{arXiv preprint arXiv:2301.03396}, 2023.

\bibitem{mildenhall2021nerf}
B.~Mildenhall, P.~P. Srinivasan, M.~Tancik, J.~T. Barron, R.~Ramamoorthi, and R.~Ng, ``Nerf: Representing scenes as neural radiance fields for view synthesis,'' \emph{Communications of the ACM}, vol.~65, no.~1, pp. 99--106, 2021.

\bibitem{ji2022eamm}
X.~Ji, H.~Zhou, K.~Wang, Q.~Wu, W.~Wu, F.~Xu, and X.~Cao, ``Eamm: One-shot emotional talking face via audio-based emotion-aware motion model,'' in \emph{ACM SIGGRAPH 2022 Conference Proceedings}, 2022, pp. 1--10.

\bibitem{tan2024flowvqtalker}
S.~Tan, B.~Ji, and Y.~Pan, ``Flowvqtalker: High-quality emotional talking face generation through normalizing flow and quantization,'' in \emph{Proceedings of the IEEE/CVF Conference on Computer Vision and Pattern Recognition}, 2024, pp. 26\,317--26\,327.

\bibitem{nagrani2017voxceleb}
A.~Nagrani, J.~S. Chung, and A.~Zisserman, ``Voxceleb: a large-scale speaker identification dataset,'' \emph{arXiv preprint arXiv:1706.08612}, 2017.

\bibitem{cao2014crema}
H.~Cao, D.~G. Cooper, M.~K. Keutmann, R.~C. Gur, A.~Nenkova, and R.~Verma, ``Crema-d: Crowd-sourced emotional multimodal actors dataset,'' \emph{IEEE transactions on affective computing}, vol.~5, no.~4, pp. 377--390, 2014.

\bibitem{jackson2014surrey}
P.~Jackson and S.~Haq, ``Surrey audio-visual expressed emotion (savee) database,'' \emph{University of Surrey: Guildford, UK}, 2014.

\bibitem{metashape}
Agisoft, ``2023. metashape. https://www.agisoft.com/.''

\bibitem{zhang2016joint}
K.~Zhang, Z.~Zhang, Z.~Li, and Y.~Qiao, ``Joint face detection and alignment using multitask cascaded convolutional networks,'' \emph{IEEE signal processing letters}, vol.~23, no.~10, pp. 1499--1503, 2016.

\bibitem{king2009dlib}
D.~E. King, ``Dlib-ml: A machine learning toolkit,'' \emph{The Journal of Machine Learning Research}, vol.~10, pp. 1755--1758, 2009.

\bibitem{danvevcek2022emoca}
R.~Dan{\v{e}}{\v{c}}ek, M.~J. Black, and T.~Bolkart, ``Emoca: Emotion driven monocular face capture and animation,'' in \emph{Proceedings of the IEEE/CVF Conference on Computer Vision and Pattern Recognition}, 2022, pp. 20\,311--20\,322.

\bibitem{DECA:Siggraph2021}
\BIBentryALTinterwordspacing
Y.~Feng, H.~Feng, M.~J. Black, and T.~Bolkart, ``Learning an animatable detailed {3D} face model from in-the-wild images,'' \emph{ACM Transactions on Graphics (ToG), Proc. SIGGRAPH}, vol.~40, no.~8, 2021. [Online]. Available: \url{https://doi.org/10.1145/3450626.3459936}
\BIBentrySTDinterwordspacing

\bibitem{filntisis2022visual}
P.~P. Filntisis, G.~Retsinas, F.~Paraperas-Papantoniou, A.~Katsamanis, A.~Roussos, and P.~Maragos, ``Visual speech-aware perceptual 3d facial expression reconstruction from videos,'' \emph{arXiv preprint arXiv:2207.11094}, 2022.

\bibitem{kingma2014adam}
D.~P. Kingma and J.~Ba, ``Adam: A method for stochastic optimization,'' \emph{arXiv preprint arXiv:1412.6980}, 2014.

\bibitem{sohn2015learning}
K.~Sohn, H.~Lee, and X.~Yan, ``Learning structured output representation using deep conditional generative models,'' \emph{Advances in neural information processing systems}, vol.~28, 2015.

\bibitem{baevski2020wav2vec}
A.~Baevski, Y.~Zhou, A.~Mohamed, and M.~Auli, ``wav2vec 2.0: A framework for self-supervised learning of speech representations,'' \emph{Advances in neural information processing systems}, vol.~33, pp. 12\,449--12\,460, 2020.

\bibitem{sahidullah2012design}
M.~Sahidullah and G.~Saha, ``Design, analysis and experimental evaluation of block based transformation in mfcc computation for speaker recognition,'' \emph{Speech communication}, vol.~54, no.~4, pp. 543--565, 2012.

\bibitem{yi2023generating}
H.~Yi, H.~Liang, Y.~Liu, Q.~Cao, Y.~Wen, T.~Bolkart, D.~Tao, and M.~J. Black, ``Generating holistic 3d human motion from speech,'' in \emph{Proceedings of the IEEE/CVF Conference on Computer Vision and Pattern Recognition}, 2023, pp. 469--480.

\bibitem{dosovitskiy2020image}
A.~Dosovitskiy, L.~Beyer, A.~Kolesnikov, D.~Weissenborn, X.~Zhai, T.~Unterthiner, M.~Dehghani, M.~Minderer, G.~Heigold, S.~Gelly \emph{et~al.}, ``An image is worth 16x16 words: Transformers for image recognition at scale,'' \emph{arXiv preprint arXiv:2010.11929}, 2020.

\bibitem{lu2021live}
Y.~Lu, J.~Chai, and X.~Cao, ``Live speech portraits: real-time photorealistic talking-head animation,'' \emph{ACM Transactions on Graphics (TOG)}, vol.~40, no.~6, pp. 1--17, 2021.

\bibitem{isola2017image}
P.~Isola, J.-Y. Zhu, T.~Zhou, and A.~A. Efros, ``Image-to-image translation with conditional adversarial networks,'' in \emph{Proceedings of the IEEE conference on computer vision and pattern recognition}, 2017, pp. 1125--1134.

\bibitem{siyao2022bailando}
L.~Siyao, W.~Yu, T.~Gu, C.~Lin, Q.~Wang, C.~Qian, C.~C. Loy, and Z.~Liu, ``Bailando: 3d dance generation by actor-critic gpt with choreographic memory,'' in \emph{Proceedings of the IEEE/CVF Conference on Computer Vision and Pattern Recognition}, 2022, pp. 11\,050--11\,059.

\bibitem{heusel2017gans}
M.~Heusel, H.~Ramsauer, T.~Unterthiner, B.~Nessler, and S.~Hochreiter, ``Gans trained by a two time-scale update rule converge to a local nash equilibrium,'' \emph{Advances in neural information processing systems}, vol.~30, 2017.

\bibitem{parmar2021cleanfid}
G.~Parmar, R.~Zhang, and J.-Y. Zhu, ``On aliased resizing and surprising subtleties in gan evaluation,'' in \emph{CVPR}, 2022.

\bibitem{zhang2018unreasonable}
R.~Zhang, P.~Isola, A.~A. Efros, E.~Shechtman, and O.~Wang, ``The unreasonable effectiveness of deep features as a perceptual metric,'' in \emph{Proceedings of the IEEE conference on computer vision and pattern recognition}, 2018, pp. 586--595.

\bibitem{wang2004image}
Z.~Wang, A.~C. Bovik, H.~R. Sheikh, and E.~P. Simoncelli, ``Image quality assessment: from error visibility to structural similarity,'' \emph{IEEE transactions on image processing}, vol.~13, no.~4, pp. 600--612, 2004.

\bibitem{narvekar2011no}
N.~D. Narvekar and L.~J. Karam, ``A no-reference image blur metric based on the cumulative probability of blur detection (cpbd),'' \emph{IEEE Transactions on Image Processing}, vol.~20, no.~9, pp. 2678--2683, 2011.

\bibitem{chen2018lip}
L.~Chen, Z.~Li, R.~K. Maddox, Z.~Duan, and C.~Xu, ``Lip movements generation at a glance,'' in \emph{Proceedings of the European conference on computer vision (ECCV)}, 2018, pp. 520--535.

\bibitem{bulat2017far}
A.~Bulat and G.~Tzimiropoulos, ``How far are we from solving the 2d \& 3d face alignment problem? (and a dataset of 230,000 3d facial landmarks),'' in \emph{International Conference on Computer Vision}, 2017.

\end{thebibliography}


 





\end{document}